\documentclass[review]{elsarticle}

\usepackage{multirow}
\usepackage{graphicx} 
\usepackage{wrapfig}
\usepackage{caption}
\usepackage{subcaption}
\usepackage{eqnarray,amsmath}
\usepackage{colortbl}
\usepackage{tabulary}
\usepackage{stmaryrd}
\usepackage{ifsym}

\usepackage{hyperref}

\journal{EXPERT SYSTEMS WITH APPLICATIONS }

\biboptions{authoryear}

\begin{document}

\begin{frontmatter}

\title{Estimation of the qualification and behavior of a contributor and aggregation of his answers in a crowdsourcing context}

\author{Constance Thierry\fnref{druid}}
\ead{constance.thierry@irisa.fr}
\author{Arnaud Martin\fnref{druid}}
\ead{arnaud.martin@irisa.fr}
\author{Jean-Christophe Dubois\fnref{druid}}
\ead{jean-christophe.dubois@irisa.fr}
\author{Yolande Le Gall\fnref{druid}}
\ead{yolande.le-gall@irisa.fr}

\fntext[druid]{{http://www-druid.irisa.fr},
            {Univ. Rennes, CNRS, IRISA, DRUID, France}}

\end{frontmatter}

\begin{frontmatter}

\begin{abstract}
Crowdsourcing is the outsourcing of tasks to a crowd of contributors on a dedicated platform.
The crowd on these platforms is very diversified and includes various profiles of contributors which generates data of uneven quality.
However, majority voting, which is the aggregating method commonly used in platforms, gives equal weight to each contribution.
To overcome this problem, we propose a method, MONITOR, which estimates the contributor's profile and aggregates the collected data by taking into account their possible imperfections thanks to the theory of belief functions.
To do so, MONITOR starts by estimating the profile of the contributor through his qualification for the task and his behavior.
Crowdsourcing campaigns have been carried out to collect the necessary data to test MONITOR on real data in order to compare it to existing approaches.
The results of the experiments show that thanks to the use of the MONITOR method, we obtain a better rate of correct answer after aggregation of the contributions compared to the majority voting.
Our contributions in this article are for the first time the proposal of a model that takes into account both the qualification of the contributor and his behavior in the estimation of his profile. For the second one, the weakening and the aggregation of the answers according to the estimated profiles.
\end{abstract}


\end{frontmatter}


\section{Introduction}

The term crowdsourcing is a neologism introduced by \cite{howe06} combining the terms crowd and outsourcing.
It consists in outsourcing tasks that cannot be performed by computers because they cannot be automated or require human expertise, to a crowd of contributors on a dedicated platform.
The diversity of tasks that can be performed leads to a diversity of existing platforms, so that different nomenclatures appear in the literature.
\cite{guittard10} and \cite{schenk12} define three categories according to the employer's expectations: routine, complexity and creativity.
In our research we only consider data from crowdsourcing platforms of routine activities.
The platforms of this type propose simple micro-tasks, which do not require any particular qualification, and are achievable in a short time and remunerated by micro-payments.
The employer can be an industrial company as introduced by \cite{howe06}, but it can also be a research laboratory. 
Indeed, \cite{xintong14} reviews the use of crowdsourcing for data mining.
The crowd on these platforms is large, with diversified profiles as shown in the demographic study of \cite{ross09}. 
Amazon mechanical Turk\footnote{https://www.mturk.com (27/10/2021)} (AMT) is a routine activity platform used all over the world. 
In order to ensure the quality of the data collected, it is possible to act upstream of the campaign by selecting the contributors, as \cite{yadav22} does. Another possibility is to consider data processing after the campaign. In this paper we deal with this second possibility by addressing the aggregation of responses and the expertise of the contributor.\\

There are several methods of aggregating responses for crowdsourcing, the article of \cite{sheng19} summarises a number of them. 
Usually the data collected during crowdsourcing campaigns are aggregated by majority voting (MV).
It is for example used by \cite{nguyen15}.
This method consists in selecting the answer given by the largest number of contributors, it has the advantage of being simple to implement. 
Unfortunately the MV is not robust against some malicious contributors.
Indeed, the contributors on crowdsourcing platforms are numerous and have very diverse profiles.
On one hand, most of them are serious about the task and have the skills required for the job and there are also people among the crowd who have a deep knowledge of the proposed field.
But, on the other hand, there are a few unscrupulous contributors who behave maliciously, responding randomly and quickly in order to complete as many tasks as possible in a short period of time to maximize the reward.
The problem is that these random responses negatively impact the aggregation of responses by MV. 
\cite{lee10} improve the MV results by pre-selecting responses where bad contributors are removed. 
This selection is possible thanks to the use of golden data, {\emph i.e}. questions whose answers are known by the employer.
\cite{khattak11} also use gold data to improve the quality of MV results.
Unfortunately it is not always possible to have golden data in crowdsourcing campaigns, therefore their use is an obstacle to the widespread use of the methods proposed by the authors.

To overcome this problem it is possible to use the Expectation-Maximization (EM) algorithm of \cite{dempster77} which allows an estimation of the contributor's profile and the aggregation of the contributions.
EM has been the subject of various research works for its application to crowdsourcing~: \cite{dawid79,whitehill09,welinder10,ipeirotis10,wang11,raykar12}.
And several experiments as those ones of \cite{whitehill09,raykar10,hung13} have shown that EM offers better results than majority voting for aggregating responses.
However, EM only takes into account the qualification of the contributor, thanks to the confusion matrix on his answers, whereas it would be interesting to also consider his behavior.

Unfortunately it is not possible to know the contributor's behavior, and current studies focus more on the contributor's expertise than on his behavior. \cite{halpin12} uses Support Vector Machines to identify contributors who answer randomly. To do so, the author considers the number of tasks performed by the contributor, the average response time, and the average of the correct answers on the gold data.
The results obtained are good unfortunately this approach requires gold data which is not applicable for all crowdsourcing campaigns. \\

Let's take the example of a photo of a bird presented to a crowd of 20 people composed of 5 ornithologists and 15 people with no proven expertise in the field.
The issues considered in this paper are therefore the following:

- How to identify the right answer when it is given by a minority of experts and the rest of the crowd chooses another answer? Giving our example, let's imagine that the 5 specialists identify the bird with certainty while the rest of the crowd chooses the wrong species without much certainty. Then it would be better to choose the expert's answer, but approaches like MV indicate otherwise. The expertise of the contributor and his certainty play a crucial role here. This leads to the following questions.

- If we allow a contributor to express his answer more finely by indicating his certainty in his answer but also by offering him the possibility of being imprecise, how can we integrate these imperfections in the aggregation of answers?
Using the previous example, a person indicating that the bird is ``possibly a black tit'' is less certain than a person stating that they are ``certain it is a black tit or a great tit''.

- Finally, the expertise of the contributor in the field is also relevant and should be considered. How can this expertise be integrated into the aggregation of responses?
And in the meantime how to evaluate it? As mentioned earlier, the response of an ornithologist carries more weight than that of a neophyte.
\\

It is in this context that MONITOR is defined.
MONITOR is the acronym for MOdelling uNcertainty and Inaccuracy on daTa from crOwdsourcing platfoRms, because this model considers answers for which the contributor can be imprecise while informing his certainty.
The imprecision characterizes the information contribution of an assertion, in this paper it is the number of selected answers for a MCQ.
The certainty indicates here the epistemic certainty which reflects the knowledge of the contributor on the considered subject.
A precise answer is generally expected from a contributor whereas we offer him the opportunity to be imprecise in case of doubt while reflecting his certainty which is not common in the literature either.  But when he has the opportunity to be imprecise the contributor is more certain of his answer. Introducing imprecision and uncertainty in MCQs and modeling it by belief functions improves the quality of the results obtained for the employer according to \cite{thierry21}.
MONITOR allows the estimation of the contributor's profile and an aggregation of the answers accordingly without using gold data.
The interest of MONITOR is that the knowledge of the contributor is studied in conjunction with his behavior, which is not the case of the traditional methods of profile estimation introduced earlier.
The model originally defined by \cite{thierry19} considered only the contributor's imprecision and response time  for the profile estimation and has evolved into the more complex model presented in this paper. 
Improvements to the model enhanced profile estimation and response aggregation.\\

The rest of this paper is organized as follows. First, the theory of belief functions is explained in section~\ref{sec:BF}.  Then, existing methods that apply this theory in a crowdsourcing context are exposed section \ref{sec:EA_bf}.
The MONITOR model is then introduced in section~\ref{sec:MONITOR}.
Since the tests are carried out on real data, the crowdsourcing campaigns carried out for the acquisition of the data are presented in section~\ref{sec:campaigns}.
Section~\ref{sec:resultats} exposes the obtained results and section~\ref{sec:ccl} concludes this paper.


\section{Belief functions}
\label{sec:BF}

The theory of belief functions is also called Dempster-Shafer theory, because it was introduced by Dempster \cite{dempster67} and formalized by Shafer \cite{shafer76}.
It is a generalization of fuzzy and probabilistic approaches and allows to model the imprecision and uncertainty of imperfect sources of information.
This theory can be used to model information coming from an expert, as \cite{yang09} does, or from a crowd in the context of crowdsourcing, as \cite{thierry19} does.\\

The section \ref{sec:mass} first introduces the modeling of uncertainty and imprecision by mass functions. 
Some operations, sometimes necessary before the aggregation of mass functions, are also presented section \ref{sec:op} before presenting different existing combination operators in section \ref{sec:agg}.
The methods used to make a decision on the mass function after aggregation are finally exposed section \ref{sec:decision}.

\subsection{Mass function}
\label{sec:mass}

We call the set of classes or hypotheses $r_i$ that are exclusive and exhaustive the frame of discernment $\Omega=\{r_0,..., r_n\}$.
The belief functions are defined on:
\begin{equation}
    2^\Omega=\{ \emptyset, \{r_0\}, \{r_1\},\{r_0 \cup r_1 \},...,\Omega\}
\end{equation}
The element $\Omega$ represents ignorance, and $\emptyset$ symbolizes openness to the world outside the frame of discernment.
In crowdsourcing, a contributor $c$ is an imperfect source, and $\Omega$ is the set of possible answer choices to a question $q$.
The mass functions $m^\Omega$ model the elementary degree of belief of the source. They are defined on $2^\Omega$ with values in $[0,1]$ and respect the normalization condition:
\begin{equation}
    \sum_{X \in 2^{\Omega}} m^\Omega(X) = 1 
    \label{eq:cond_norma}
\end{equation}

The higher the mass $m^\Omega(X)$, the stronger the belief.
When $m^\Omega(\emptyset)=0$ it means that an opening to the world out of the frame of discernment is not possible, one says then to be in closed world.
Moreover, for $m^\Omega(\emptyset)>0$, the function is said to be non-dogmatic.

An element $X \in 2^\Omega$ such that $m^\Omega(X)>0$ is called focal element, and the meeting of the focal elements constitutes the kernel.
If only the singletons of $\Omega$ are focal elements then $m^\Omega$ is a probability, the function is then called Bayesian mass function, it is a specific mass function but there are others.

\paragraph{Categorical mass function} 
When the source is absolutely certain of its answer, all belief is given to $X \in 2^\Omega$. 
The answer $X$ can be imprecise in case it is a meeting of classes belonging to the frame of discernment. 
\begin{equation}
    m^\Omega(X) = 1, X \in 2^\Omega
\end{equation}
If $X$ is a singleton $r_i \in \Omega$, then not only is the answer completely certain, but it is also precise.
In the specific case where $m^\Omega(\Omega)=1$, the mass function reflects complete ignorance on the part of the information source.

\paragraph{Simple support mass functions ($X^{\omega}$)}
This mass function reflects an uncertain and imprecise response from the information source.
\begin{eqnarray} 
    \left \{
        \begin{array}{l}
            m^{\Omega}(X) = \omega   \mbox{ avec }  X \in 2^{\Omega} \setminus \Omega, \omega  \in [0,1]\\
            m^{\Omega}(\Omega) = 1 - \omega  \\
            m^{\Omega}(Y) = 0, \mbox{ }  \forall  Y \in 2^{\Omega} \setminus \{X, \Omega \}
        \end{array}
    \right.
\end{eqnarray}
The source has an uncertain knowledge because it partially believes in $X$ but not totally since a non-zero mass is present on $\Omega$.

For example, a contributor is shown a picture of a bird and asked to identify it according to the answer set: $\Omega = \{crow, raven, eagle\}$ and to indicate his certainty in his answer on a scale between 0 (not at all certain) and 1 (totally certain). 
The contributor hesitates between crow and raven and chooses to select these two propositions and indicates a certainty value of 0.7 because he is rather certain but not totally. 
The answer $X=\{crow, raven \}$ is imprecise and the resulting simple support mass function is $m^\Omega(\{crow, raven \})=0.7$ and $m^\Omega(\Omega)=0.3$.

\subsection{Operations on belief functions}
\label{sec:op}

In this section, methods to facilitate the combination of belief functions are discussed. 
From now on we denote $m_{cq}^\Omega$ the mass function associated with the contributor $c$ for the question $q$ and the frame of discernment $\Omega$.

\paragraph{Discounting}
The discounting coefficient $\alpha_c \in [0,1]$ models the confidence in the source $c$.
The discounting of a mass function is defined as follows:
\begin{eqnarray} 
    \left \{
        \begin{array}{l}
            m_{cq}^{\Omega, \alpha}(X) = \alpha_c m_{cq}^{\Omega}(X), \mbox{ } \forall  X \in 2^{\Omega} \setminus \Omega\\
            m_{cq}^{\Omega, \alpha}(\Omega) = 1 - \alpha_c(1 - m_{cq}^{\Omega}(\Omega)) \\
        \end{array}
    \right.
    \label{eq:affaiblissement}
\end{eqnarray}
The larger this coefficient is, the more reliable the source is considered to be.
If $\alpha_c=0$ the source is absolutely unreliable and the totality of the mass is transferred to the ignorance $\Omega$.
The discounting decreases the conflict resulting from the combination of the belief functions. 

\paragraph{Jousselme distance}
This distance is defined by \cite{jousselme01} to estimate the proximity between two mass functions.
Let two mass functions $m_1^\Omega$ and $m_2^\Omega$ have the same frame of discernment $\Omega$, the distance is given by:
\begin{equation}
    d_J(m_1^\Omega,m_2^\Omega) = \sqrt{\frac{1}{2}(m_1^\Omega - m_2^\Omega)^T \underline{\underline{D}} (m_1^\Omega - m_2^\Omega)} 
    \label{eq:dJ}
\end{equation}

\begin{eqnarray} 
    \underline{\underline{D}}(X,Y) = \left \{
    \begin{array}{l}
        1 \mbox{ if } X = Y = \emptyset \\
        \frac{|X \cap Y|}{|X \cup Y|} \forall X,Y \in 2^\Omega \\
    \end{array}
    \right.
    \label{eq:Jac}
\end{eqnarray}
In equation \eqref{eq:dJ}, $\underline{D}$ is a matrix of size $2^\Omega \times 2^\Omega$ based on Jaccard dissimilarity given by equation \eqref{eq:Jac}.  
The higher the distance $d_J$, the more different the mass functions are, and reciprocally, the lower it is the more similar they are.
The interest of this distance is that it takes into account the cardinality of the focal elements of the two mass functions.

\paragraph{Canonical mass function decomposition}
A non-dogmatic mass function whose focal elements $X_i \subset \Omega$ are distinct can decompose in a unique way:
\begin{equation}
    m = \displaystyle \bigcap_{X \subset \Omega} X^{\omega(X)}
    \label{eq:decompCanon}
\end{equation}
In equation \ref{eq:decompCanon}, $X^{\omega(X)}$ is a simple support mass function whose focal element $X$ has mass $\omega(X) \in [0,1]$.
The operator used for combining $X^{\omega(X)}$ to find $m$ is the conjunctive combination operator described in the next section.

\subsection{Aggregation}
\label{sec:agg}

For information fusion, the sources all report on the same frame of discernment $\Omega$ and the same topic. 
The following notations are used: a crowd composed of a total of $K$ contributors ($c$) answers a question ($q$) according to the set of possible answer $\Omega$. 
For all equations defined in this section $X \in 2^\Omega$. It is possible to average mass functions:
\begin{equation}
    m_{Avg}(X) = \frac{1}{K} \sum_{c=1}^K m_{cq}^\Omega(X)
    \label{eq:mMoy}
\end{equation}
This simplistic combination of mass functions allows us to remain in a closed world.
Many other rules of combination exist in the theory of belief functions, this section lists some conjunctive rules, \cite{martin19} presents more.
Conjunctive combination requires the sources to be reliable, distinct and cognitively independent.

\paragraph{Conjunctive rule}
This operator reduces the imprecision on the focal elements and increases the belief on the concordant elements. 
\begin{equation}
    m_{Conj}^\Omega(X) =\left( \bigcap_{c=1}^K m_{cq}^\Omega \right)(X)  =  \sum_{Y_1 \cap \ldots \cap Y_K = X}  \prod_{c=1}^K m_{cq}^\Omega(Y_c) 
    \label{eq:mConjonctive}
\end{equation}
The mass $m_{Conj}^\Omega(\emptyset)$ represents the global conflict of the combination.
Other operators allow to remain in a closed world after combination, for example the normalized conjunctive rule of Dempster or the one of \cite{yager87}.

\paragraph{Dempster rule}
This operator allows a fair distribution of the conflict \linebreak $k=m_{Conj}^\Omega(\emptyset)$ on the focal elements. 
\begin{equation}
    m_D^\Omega(X) = \frac{1}{1-k}m_{Conj}^\Omega(X)
    \label{eq:mConjNorm}
\end{equation}

\paragraph{LNS rule}
Sometimes, the conjunctive rule does not allow to obtain decidable results.
It is in particular the case when the number of sources to combine is high or are not all reliable, as for human sources.
The LNS rule given by equation \eqref{eq:LNS} and proposed by \cite{zhou17} presents the interest to decrease the constraint of reliability of the sources.
Indeed, the LNS rule requires for its application that the sources are cognitively independent and mostly reliable. 
For this rule, the more consistent a source is with others, the more reliable it is.
A canonical decomposition of the mass functions $m_{cq}^\Omega$ is performed for each contributor $c$ for the same question $q$ in order to obtain the set of simple support mass functions $\{X_{l}^{\omega_{cq}}, X_l \subset \Omega\}$.
The mass functions ${X_l^{\omega_{cq}}}$ are then grouped into $L$ clusters, $L$ being the distinct number of $X_l$.
Each cluster consists of a number $s_l$ of single supported mass functions.
\begin{equation}
    \displaystyle m_{LNS}^\Omega = \bigcap_{l=1,...,L} (X_l)^{ \displaystyle 1 - \alpha_l\left(1 - \prod_{c=1}^{s_l} \omega_{cq}\right)}
    \label{eq:LNS}
\end{equation}
\begin{equation}
     \alpha_l = \frac{s_l}{\displaystyle \sum_{i=1}^L s_i}
\end{equation}
In equation \eqref{eq:LNS}, $\alpha_l$ is the average number of simple support mass functions present in cluster $l$ over the total number of mass functions generated by the decomposition. 

Having presented combination rules for the theory of belief functions, we now turn to decision making.

\subsection{Decision}
\label{sec:decision}

Combining the sources of information yields the mass $m_{Comb}$ which is the result of aggregating the mass functions.
There are different strategies for making decisions about $m_{Comb}$ in the theory of belief functions. 
Here we present the decision on the maximum of pignistic probability and the Jousselme distance.

\paragraph{Maximum of pignistic probability}
The credal level, which consists of the modeling and manipulation of information, differs from the pignistic level, which allows decision making.
The pignistic probability $betP$ is defined by \cite{smets90} as:
\begin{equation}
 \displaystyle betP(X) = \sum_{Y \in 2^{\Omega}, Y \neq \emptyset} \frac{|X \cap Y|}{|Y|} \frac{m^\Omega(Y)}{1 - m^\Omega(\emptyset)}
 \label{eq:betP}
\end{equation}
The maximum probability is obtained for $r_d$ such that:
\begin{equation}
    betP(r_d) = \displaystyle \max_{r_i \in \Omega}betP(r_i)
\end{equation}

\paragraph{Decision distance based}
\cite{essaid14} propose to make a decision using the Jousselme distance. To do this, $d_J$ is computed between the mass resulting from the combination of the information $m_{Comb}$ and a categorical mass function $m_X$; $X \in 2^\Omega$ is the unique focal element of this function:
\begin{equation}
    X_d =\displaystyle \arg \min_{X \in 2^\Omega} d_J(m_{Comb},m_X)
    \label{eq:decisionJ}
\end{equation}
The element $X_d \in 2^\Omega$ chosen is the one that minimizes the distance between the combination of belief functions and the categorical mass function.
This decision solution is interesting because $X$ is not necessarily a singleton. \\

The following section presents the state of the art of the use of belief function theory in a crowdsourcing context.


\section{Related work}
\label{sec:EA_bf}

In crowdsourcing platforms, belief functions are used for estimating the contributor's profile or aggregating responses.
The theory of belief functions also makes it possible to model the imprecision of contributions, which is a real advantage for crowdsourcing.
Indeed, \cite{smets97} emits the hypothesis that the more imprecise an individual is, the more certain he is and in reverse, the more precise he is the less certain he is.
This hypothesis is verified by \cite{thierry21} in a context of crowdsourcing.
This is why allowing the contributor to be imprecise is an advantage for the employer since the responses of the crowd are then more certain.
To our knowledge, there are few elements in the literature that deal with the theory of belief functions applied to crowdsourcing. 
However, among those that do exist, we differentiate between approaches for aggregating responses (section \ref{sec:AnsAg}), estimating the contributor's profile (section \ref{sec:BF_profil}) and performing these two operations jointly (section \ref{sec:BF_profil_agg}).

\subsection{Answer aggregation}
\label{sec:AnsAg}

The CASCAD method for modeling and aggregating contributions from crowdsourcing platforms has been defined by \cite{koulougli16}.
In order to apply CASCAD, the contributor must perform a qualification test before the campaign to determine his expertise according to the following scale. 
During the campaign, the participant selects one to several answers and assigns to his contributions his degree of uncertainty which allows to build mass functions.
These functions are weakened by a $\alpha$ coefficient related to the previously established expertise of the contributor, after which the responses are aggregated by Dempster's combination operator (equation~\eqref{eq:mConjNorm}).
CASCAD is compared to the majority vote and the EM algorithm of \cite{dawid79}, generated data are used for testing.
CASCAD obtains a higher correct response rate than EM and MV when three focal elements are used. On the other hand, this method is more expensive in terms of execution time and memory. 

\subsection{Profile estimation}
\label{sec:BF_profil}
Two approaches exist for the estimation of the profile: the method of \cite{dubois19} which requires gold data and that of \cite{rjab16} which is free of it.  

\cite{dubois19} estimate the expertise of the contributors in order to distinguish the ``experts'' $E$ from the ``non-experts'' $NE$ according to the frame of discernment $\Omega = \{E, NE \}$.
The authors build a graph oriented on the expected answers thanks to the gold data, another graph is defined according to the answers given by the contributor.
In order to compare the reference graph to the contributor's graph, mass functions are computed for each node:
\begin{itemize}
    \item $m_1^\Omega$ compare the position of a node between the two graphs.
    \item $m_2^\Omega$ measures the proportion of nodes with the same distance to the starting point of the graph as the considered node.
    \item $m_3^\Omega$ and $m_4^\Omega$ measure the inversion errors between the previous and following nodes of a given node.
\end{itemize}
The mass function for the whole graph is computed from the average of the mass functions $m_1^\Omega$, $m_2^\Omega$, $m_3^\Omega$ and $m_4^\Omega$ which allows to estimate the expertise of the contributor.   
In order to test their model, the authors used real data obtained through crowdsourcing campaigns to evaluate the quality of sound recordings.
One of the campaigns was performed by a crowd of contributors living in Asia and another by contributors residing in the United States of America.
The degrees of expertise calculated for these two campaigns are compared and it appears that the expertise of the American contributors is higher than that of the Asian ones.
This phenomenon is explained in the article as resulting from the cultural differences of the two continents. 
However, it is not always possible for the employer to have golden data, so the approach of \cite{rjab16} which is not subject to this constraint is interesting.

In order to identify contributors who are experts in their domain without resorting to gold data, \cite{rjab16} computes a degree of precision $DP_c$, equation \eqref{eq:DPc}, and a degree of correctness $DE_c$, equation \eqref{eq:DEc}, on the answer.
Let $E_C$ be the set of contributors, $E_{Q_c}$ the set of questions answered by a contributor $c$ and $\Omega_q$ the frame of discernment associated with question $q$. 
\begin{eqnarray}
 \left \{
 \begin{array}{l}
  \displaystyle DE_c = 1 - \frac{1}{|E_{Q_c}|} \sum_{q \in E_{Q_c}} d_J(m_c^{\Omega_q} , m_{E_C|c}^{\Omega_q}  ) \\
  \displaystyle m_{E_C|c}^{\Omega_q}(X) = \frac{1}{|E_C|-1} \sum_{j \in E_C|c} m_j^{\Omega_q}(X)
 \end{array}
 \right.
 \label{eq:DEc}
\end{eqnarray}
The degree of correctness $DE_c$ reflects the overall accuracy of contributor $c$ answers compared to the aggregated answers of the rest of the crowd.
However, this degree is only relevant if the majority of the crowd is correct.
The equation is based on $d_J$ the distance of \cite{jousselme01}.
\begin{eqnarray}
  \left \{
  \begin{array}{l}
  \displaystyle DP_c =  \frac{1}{|E_{Q_c}|} \sum_{q \in E_{Q_c}} \delta_c^{\Omega_q}\\
  \displaystyle \delta_c^{\Omega_q} = 1 - \sum_{X \in 2^{\Omega_q}} m_c^{\Omega_q}(X) \frac{log_2(|X|)}{log_2(|\Omega_q|)}
 \end{array}
 \right.
 \label{eq:DPc}
\end{eqnarray}
The degree of precision $DP_c$ measures the dispersion of the responses weighted by their mass. 
The global degree of expertise of the contributor $DG_{c}$ calculated by the authors is the sum of $DE_c$ and $DP_c$ weighted by a coefficient $\beta_{c} \in [0,1]$:  
\begin{equation}
 DG_{c} = \beta_{c} DE_{c} + (1-\beta_{c}) DP_{c} 
  \label{eq:rjab}
\end{equation}
The study of~\cite{rjab16} proposes a comparison with a probabilistic approach measuring the expertise of a contributor. Generated data are used for the experiments.
The results obtained by the authors show that the calculation of $DG_{c}$ is more relevant for the evaluation of experts than the probabilistic approach. 

\subsection{Profile estimation and answers aggregation}
\label{sec:BF_profil_agg}
The method of \cite{abassi18}, named CGS-BLA\footnote{Clustering approach of the Gold Standards based Belief} allows the estimation of the contributor's profile and the aggregation of his responses. 
To do so, the authors define three profiles: the ``Expert'', the ``Good contributor'' and the ``Bad contributor''.
To identify the profile of the contributor $c$, three measures of the accuracy of the answer are computed:
\begin{itemize}
    \item the response rate of $c$ in agreement with the gold data 
    \item the response rate of $c$ in agreement with the answers aggregated by MV
    \item the proportion of responses from the rest of the crowd that are similar to $c$'s response
\end{itemize}
Once these measures obtained, the classification algorithm k-mean is applied with k=3 to classify the contributors.
The responses of $c$ are modeled by belief functions and weakened by a value $\alpha_c$ according to the contributor's profile.
If $c$ is an expert $\alpha_c=1$, the contributor's answer is unchanged, conversely, for a ``Bad contributor'' $\alpha_c=0$, no credit is given to the contributor's answer and all belief is equated to ignorance.
A ``Good contributor'' is given a discounting equal to its rate of correct answers on the gold data.
The mass functions are then combined by the CWAC operator of \cite{lefevre13} for each question and the pignistic probabilities of each answer are calculated for decision making.
CGS-BLA offers better results in terms of accuracy compared to MV. 

\cite{thierry19} propose a first version of MONITOR for the estimation of the contributor's profile from his qualification and his behavior. 
In this version the qualification consists in the ability of the contributor to be imprecise and the behavior is modeled by the response time taken by the contributor to complete the campaign. 
The tests are performed on real data coming from crowdsourcing campaigns that consist in rating audio recordings.
The contributor is asked to rate the sound quality of the recording from 1 to 5 and if he/she hesitates between two successive ratings, for example ``3 and 4'', he/she can choose both.
However, it is not possible for the contributor to choose a score of 1 and a score higher than 2 in the same answer.
This dataset is particular because it introduces a notion of order in the proposed answers which is not always the case in crowdsourcing and can impact the contributor's answer.
When testing their version of MONITOR on these data, \cite{thierry19} obtain better results than the MV.
However, no comparison is made with EM.

\section{MONITOR}
\label{sec:MONITOR}

\begin{figure}
    \centering
    \includegraphics[height=8.5cm]{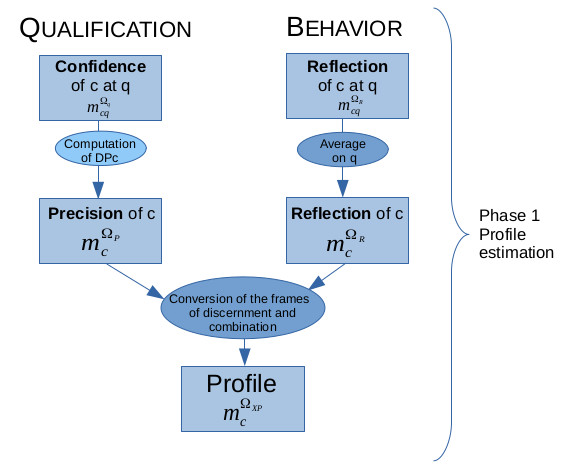}
    \caption{Scheme of MONITOR of \cite{thierry19}}
    \label{fig:MONITOR_ICTAI}
\end{figure}

\begin{figure}
    \centering
    \includegraphics[height=8.5cm]{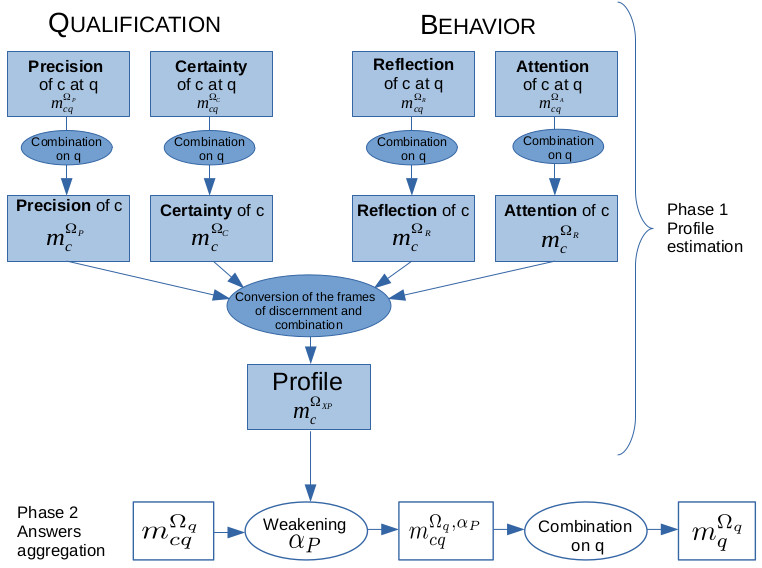}
    \caption{Scheme of the new MONITOR}
    \label{fig:MONITOR}
\end{figure}

Most elements of the state of the art simply assess the qualification of the contributor for the task, i.e., whether he/she has the required skills or is a domain expert.
MONITOR goes further by estimating the profile of the contributor not only by his qualification for the task but also by his behavior.
The model described by \cite{thierry19} considered only the estimation of the contributor's profile. 
To do so, only his ability to be precise in his contributions and his reflection were considered as shown in figure~\ref{fig:MONITOR_ICTAI}.
In the current modeling profile estimation is the first phase of the model, as shown in figure~\ref{fig:MONITOR}, and the second phase is the aggregation of contributions based on the estimated contributor profiles.
New elements have been added to the model for profile estimation.
This section presents the assumptions of the model and the two phases that compose it: profile estimation and response aggregation.

\subsection{Assumptions of the model}

The data used by MONITOR come from crowdsourcing campaigns consisting exclusively of closed questions.
Thus, for a question $q$, the set of answers associated to $q$ compose the frame of discernment $\Omega_q=\{r_1,...,r_M\}$. 

When he performs the task, the contributor $c$ has the possibility to be imprecise by selecting several answers $r_i$ which constitutes his contribution $X \in 2^{\Omega_q}$ to which he must associate a degree of certainty $\omega_{cq}$.
The answers collected in this way are therefore potentially imprecise with a variable level of certainty.
Moreover there is no interaction between the contributors so that the answers are independent.
The contribution is modeled by a single support mass function: 
\begin{equation} 
    \left \{
    \begin{array}{l}
        \displaystyle  m_{cq}^{\Omega_q}(X) = \omega_{cq}   \mbox{ avec }  X \in 2^{\Omega_q} \setminus \Omega\\
        \displaystyle  m_{cq}^{\Omega_q}(\Omega_q) = 1 - \omega_{cq} 
    \end{array}
    \right.
    \label{eq:X^w}
\end{equation}
This modeling characterizes the fact that $c$ partially believes in its answer.

The contributor's response time to a question is also saved, as it is part of the data used by MONITOR.
Crowdsourcing campaigns also include one or more attention questions $q_A$ to ensure that the contributor is serious. 
For the attention questions, the contributor is asked again a question that he has already answered and he has to fill in the same answers as the one given previously. 

\subsection{MONITOR Phase 1: Profile estimation}

For the estimation of the contributor's profile, MONITOR considers his qualification for the task and his behavior as shown in figure \ref{fig:MONITOR}.

\subsubsection{Qualification}

In MONITOR the qualification defines an assessment of the professional value of a person according to the certainty and precision of his/her answers.
To estimate the qualification of a contributor the model presented by \cite{thierry19} considers only the precision of the person. The evolution of the model led us to take into account not only the accuracy of the contributor but his certainty.

\paragraph{Precision $\Omega_\mathcal{P} = \{\mathcal{P}, \mathcal{IP}\}$}
Precision in the model qualifies the ability of the contributor to be precise $\mathcal{P}$ or imprecise $\mathcal{IP}$ in his answers. 
The more qualified a contributor is for the task, the more precise his answers are.

The mass function defined on $2^{\Omega_\mathcal{P}}$ for a contributor $c$ filling in the answer $X \in 2^{\Omega_q} \setminus \emptyset$ to the question $q$, is given by the following equations:
\begin{equation}
    \omega_{cq}^\mathcal{P} = \frac{\log_2|X|}{\log_2(imp_{MAX})}
    \label{eq:wI}
\end{equation}
\begin{eqnarray}
 \left \{
 \begin{array}{l}
  \displaystyle  m_{cq}^{\Omega_\mathcal{P}}(\mathcal{P}) = \alpha^\mathcal{P} * (1 - \omega_{cq}^\mathcal{P} )\\
  \displaystyle m_{cq}^{\Omega_\mathcal{P}}(\mathcal{IP}) = \alpha^\mathcal{P} * \omega_{cq}^\mathcal{P}\\
  \displaystyle  m_{cq}^{\Omega_\mathcal{P}}(\Omega_\mathcal{P}) = 1 - \alpha^\mathcal{P}
 \end{array}
 \right.
 \label{eq:mI}
\end{eqnarray}
$\alpha^\mathcal{P} \in [0,1]$ in equation \eqref{eq:mI} is a discounting coefficient and $\omega_{cq}^\mathcal{P}$ is the mass associated with the elements of $\Omega_\mathcal{P}$.
The computation of the mass $\omega_{cq}^\mathcal{P}$ is based on the $DP_c$ degree of precision of \cite{rjab16} because in the version of \cite{thierry19} of MONITOR it is this $DP_c$ degree that is used.
Unlike \cite{rjab16} and \cite{thierry19} who use $\log_2|\Omega_q|$, we have $\log_2(imp_{MAX})$ with $imp_{MAX}$ the maximum imprecision the employer allows the contributor ($|X| < imp_{MAX} \leq |\Omega_q|$).
The more imprecise the contributor is the higher $\omega_{cq}^\mathcal{P}$ is.
For a precise answer, $X$ is a singleton and the mass $\omega_{cq}^\mathcal{P}=0$ thanks to the logarithm function and $m_{cq}^{\Omega_\mathcal{P}}$ becomes a simple supported mass function with $\mathcal{P}$ as focal element. 
Respectively, if $X=\Omega_q$, $m_{cq}^{\Omega_\mathcal{P}}$ is a simple supported mass function with $\mathcal{IP}$ as focal element. 

Unlike the degree $DP_c$ of \cite{rjab16}, the quotient of logarithms of equation \eqref{eq:wI} is not multiplied by $m_{cq}^{\Omega_q}(X)$. 
Indeed, this can eventually bias the computation of the mass on imprecision, for example if the contributor $c$ is very imprecise and selects all the answers that are proposed to him, then $X=\Omega_q$.
Let's assume that this contributor remains very uncertain about his answer despite everything, so that $m_{cq}^{\Omega_q}(X)=0$ then $m_{cq}^{\Omega_q}(X)=0$ which would mean that $c$ is totally precise when it is not.
So $m_{cq}^{\Omega_q}(X)$ can have a negative impact on the contributor's precision estimate.

\paragraph{Certainty $\Omega_\mathcal{C} = \{\mathcal{C}, \mathcal{UC} \}$}
We consider that the more qualified a contributor is, the more certain ($\mathcal{C}$) he/she is about his/her answers.
On the contrary, a less qualified contributor is more uncertain ($\mathcal{UC}$). 
When the contributor $c$ answers the question $q$, he informs his certainty $\omega_{cq}$ about the correctness of his contribution.
The value of $\omega_{cq}$ is included in the interval $[\omega_{MIN}, \omega_{MAX}]$, with $\omega_{MIN} < \omega_{MAX}$ to obtain a value of $\omega_{cq}^\mathcal{C}$ between 0 and 1. 
The value $\omega_{MIN}$ means that $c$ is not certain of his answer. On the contrary, for $\omega_{cq}=\omega_{MAX}$ the contributor is absolutely certain that his answer is correct.
The mass function associated to the certainty on $2^{\Omega_\mathcal{C}}$ is:
\begin{equation}
    \omega_{cq}^\mathcal{C} = \frac{\omega_{cq} - \omega_{MIN}}{\omega_{MAX} - \omega_{MIN}}
    \label{eq:wC}
\end{equation}

\begin{eqnarray}
    \left \{
    \begin{array}{l}
        \displaystyle  m_{cq}^{\Omega_\mathcal{C}}(\mathcal{C}) =\alpha^\mathcal{C} * \omega_{cq}^\mathcal{C}\\
        \displaystyle m_{cq}^{\Omega_\mathcal{C}}(\mathcal{UC}) =\alpha^\mathcal{C} * (1 - \omega_{cq}^\mathcal{C}) \\
        \displaystyle  m_{cq}^{\Omega_\mathcal{C}}(\Omega_\mathcal{C}) = 1 - \alpha^\mathcal{C}\\
    \end{array}
    \right.
    \label{eq:mC}
\end{eqnarray}
In equation \eqref{eq:mC}, $\alpha^\mathcal{C} \in [0,1]$ is the discounting coefficient of the function $m_{cq}^{\Omega_\mathcal{C}}$.
According to equation \eqref{eq:wC}, $\omega_{cq}^\mathcal{C}$ grows with $\omega_{cq}$ reinforcing the belief that contributor $c$ is certain $\mathcal{C}$.
For $\omega_{cq}=\omega_{MIN}$, $\omega_{cq}^\mathcal{C}=0$ and $m_{cq}^{\Omega_\mathcal{C}}(\mathcal{C})=0$, making $m_{cq}^{\Omega_\mathcal{C}}$ a focal element simple support mass function $\mathcal{UC}$.

In parallel to the qualification of the contributor, his behavior is estimated in order to determine his profile. 
The following section introduces this model.

\subsubsection{Behavior}

 \cite{mehmood16} argued that an employee's personality as defined by the Big Five model has a significant impact on the individual's job performance. 
This is consistent with the work of \cite{kazai12}. The authors show that Openness to Experience and Conscientiousness of the contributor are strongly related to the accuracy of his answers in crowdsourcing platforms.

MONITOR estimates the Conscientiousness of the contributor through their behavior by modeling their reflection and attention, whereas the previous version of the model only consider the reflection.
A person who takes time to reflect is conscientious in completing the task and will provide answers that the employer can trust.
In contrast, someone who responds quickly, and therefore with very little time to think, may turn out to be a poor contributor as well as an expert.
The bad contributor answers randomly and therefore hastily and without thinking. On the other hand, the expert responds quickly compared to the rest of the crowd because of his better knowledge of the domain.
The estimation of attention is necessary to differentiate these two profiles, because the bad contributor is not attentive to the task unlike the expert.

\paragraph{Reflection $\Omega_\mathcal{R} = \{\mathcal{R}, \mathcal{NR}\}$}
\cite{gadiraju15} highlight the correlation between the rate of good response of the contributors and their response time.
Moreover \cite{difallah12} specify that the response time is a good indicator of random contributions.
For these reasons, MONITOR's reflection modeling relies on the use of the $T_{cq}$ response time of the $c$ contributor to the $q$ question.

We assume that question $q$ requires a minimum response time $T_{0q}$. 
The value of $T_{0q}$ is estimated differently depending on the crowdsourcing campaign.
The way $T_{0q}$ is calculated is notably different in the work of \cite{thierry19} because the crowdsourcing campaign consisted of listening to sound recordings so the recording time was used.
Moreover the method used to calculate $m_{cq}^{\Omega_\mathcal{R}}$ is also different and is based on an algorithm given in the article.

If $T_{cq} < T_{0q} $, this indicates that the contributor did not take the minimum time required to think about the question. He is then considered as not reflected in his contribution ($\mathcal{NR}$).
On the contrary, for $T_{cq} \geq T_{0q}$ the contributor is reflected ($\mathcal{R}$). 
In the actual version of MONITOR the mass function modeling the reflection on $2^{\Omega_\mathcal{R}}$ is defined by: 
\begin{equation}
    \left \{
    \begin{array}{l}
        \displaystyle x = T_{cq}-T_{0q}\\
        \displaystyle \omega_{cq}^\mathcal{R} = \displaystyle  \frac{\arctan{(x)}}{\pi} + \frac{1}{2}
    \end{array}
 \right.
 \label{eq:wR}
\end{equation}
\begin{eqnarray}
 \left \{
 \begin{array}{l}
  \displaystyle  m_{cq}^{\Omega_\mathcal{R}}(\mathcal{R}) = \alpha^\mathcal{R} \omega_{cq}^\mathcal{R}\\
  \displaystyle m_{cq}^{\Omega_\mathcal{R}}(\mathcal{NR}) = \alpha^{\mathcal{R}} (1 - \omega_{cq}^\mathcal{R})\\
  \displaystyle  m_{cq}^{\Omega_\mathcal{R}}(\Omega_\mathcal{R}) = 1 - \alpha^\mathcal{R}
 \end{array}
 \right.
 \label{eq:mR}
\end{eqnarray}
In equation \eqref{eq:mR}, $\alpha_\mathcal{R} \in [0,1]$ is a discounting coefficient.
The value $x \in [-T_{0q}, +\infty[$ of equation \eqref{eq:wR} is not in $[0,1]$ because it is negative if $T_{cq} < T_{0q}$ and greater than 1 for $T_{cq} > (T_{0q}+1)$.  
The arctangent function is used in the calculation of $ \omega_{cq}^\mathcal{R}$ in order to reduce the values in the interval $[0,1]$.
This function is chosen because the response time of a contributor to a question can vary a lot within the crowd for the same question. However, when this response time is very strongly higher than $T_{0q}$ the mass on the $R$ element must be close to 1 which is possible thanks to the arctangent asymptotes.

\paragraph{Attention $\Omega_\mathcal{A} = \{\mathcal{A}, \mathcal{NA}\}$}
Attention questions are asked to the contributor during the crowdsourcing campaigns in order to ensure their seriousness.
MONITOR uses specific attention questions $q_A$  to ensure that the contributor is serious. 
This involves asking the contributor the question $q$ that precedes $q_A$ again and asking him to fill in the exact same answers. 
If the contributor is attentive ($\mathcal{A}$), he remembers his previous answers. If not ($\mathcal{NA}$), they are different.
We estimate the attention of the contributor by computing the proximity of the original answer to the answer of the attention question by a distance.

The answer $X \in 2^{\Omega_q}$ of a contributor $c$ to question $q$ is modeled by a simple support mass function $X^{\omega_{cq}}$ given by equation \eqref{eq:X^w}.
The answer $Y \in 2^{\Omega_q}$ to the attention question $q_A$, which repeats the question $q$, is similarly modeled by $Y^{\theta_{cq}}$, with $\theta_{cq} \in [0,1]$ the certainty given at $q_A$.
The mass function associated with attention is given by:
\begin{equation}
  \displaystyle \omega_{cq}^\mathcal{A} = d_J( X^{\omega_{cq}} ,Y^{\theta_{cq}})
  \label{eq:wA}
\end{equation}
\begin{eqnarray}
 \left \{
 \begin{array}{l}
  \displaystyle  m_{cq_\mathcal{A}}^{\Omega_\mathcal{A}}(\mathcal{A}) = \alpha^\mathcal{A} * (1 - \omega_{cq}^\mathcal{A})\\
  \displaystyle m_{cq_\mathcal{A}}^{\Omega_\mathcal{A}}(\mathcal{NA}) = \alpha^\mathcal{A} * \omega_{cq}^\mathcal{A}\\
  \displaystyle  m_{cq_\mathcal{A}}^{\Omega_\mathcal{A}}(\Omega_\mathcal{A}) = 1 - \alpha^\mathcal{A}
 \end{array}
 \right.
 \label{eq:mA}
\end{eqnarray}
In equation \eqref{eq:wA}, $d_J$ is the distance of \cite{jousselme01} between $X^{\omega_{cq}}$ and $Y^{\theta_{cq}}$.
We chose this metric because it takes into account the cardinality of the answer (imprecision) and the values of the masses (certainty). 
If $X=Y$ and $\omega = \theta$, then the contributions are identical and the distance is zero. $\omega_{cq}^A=1$ and $m_{cq_\mathcal{A}}^{\Omega_\mathcal{A}}$ becomes a simple support mass function of focal element $\mathcal{A}$.
Thus, using $d_J$, the closer the answer is and certainty given to the attention question is to that given to the original question, the greater the mass given to $\mathcal{A}$.

The mass functions $m_{cq}^{\Omega_\mathcal{P}}$, $m_{cq}^{\Omega_\mathcal{C}}$, $m_{cq}^{\Omega_\mathcal{R}}$, and $m_{cq_A}^{\Omega_\mathcal{A}}$ are computed for each question $q$ answered by contributor $c$.
These functions are then combined on their respective frame of discernment and on the whole campaign in order to obtain for $c$ its: precision $m_{c}^{\Omega_\mathcal{P}}$, certainty $m_{c}^{\Omega_\mathcal{C}}$, reflection $m_{c}^{\Omega_\mathcal{R}}$ and attention $m_{c}^{\Omega_\mathcal{A}}$.
Once these four elements are obtained, it is possible to estimate the profile of the contributor as shown in Figure \ref{fig:MONITOR} page \pageref{fig:MONITOR}.

\subsubsection{Profile}

\cite{thierry19} define four contributor profiles: Expert, Categarical, Fuzzy and Spammer, their definition is recalled table \ref{tab:old_profiles}.
These profiles have evolved to constitute the frame of discernment $\Omega_\mathcal{XP}=\{Expert, Good, \linebreak Average, Bad\}$ summarized in the table \ref{tab:new_profiles}.
We chose these profiles because we believe that a contributor can be excellent, correct, or poor in his or her task performance.
A contributor who excels at the task is an expert, a contributor whose answers are correct is good.
And finally those whose contributions are mediocre are average or bad contributors, differentiated by their seriousness in their work.
Indeed, we think it is a problem to penalize in the same way a contributor who gives bad answers but is conscientious and a contributor who gives bad answers because he is not attentive.
On the contrary, one could consider helping the average contributor to improve in the realization of the task.
This is why the distinction is made between the average contributor and the bad one.

\begin{table}[]
    \centering
    \begin{tabular}{|r|l|}
        \hline
        Expert & Imprecise, Not Reflected \\
        \hline
        Categorical & Precise, Reflected \\
        \hline
        Fuzzy & Imprecise, Reflected \\
        \hline
        Spammer & Precise, Not Reflected \\
        \hline
    \end{tabular}
    \caption{Profiles defined in \cite{thierry19}}
    \label{tab:old_profiles}
    \begin{tabular}{|r|l|}
        \hline
        Expert & Precise, Certain, Not Reflected, Attentive \\
        \hline
        Good & Imprecise, Certain, Reflected, Attentive\\
        \hline
        Average & Imprecise, Uncertain, Reflected, Attentive\\
        \hline
        Bad & Precise, Certain, Not Reflected, Not Attentive\\
        \hline
    \end{tabular}
    \caption{MONITOR profiles}
    \label{tab:new_profiles}
\end{table}

\paragraph{The expert} This contributor has excellent knowledge of the task area. He is therefore more qualified than the rest of the crowd.
This superior qualification is characterized by precise and certain answers.
He completes the task faster than the majority of the contributors because his answers are instinctive and therefore not reflexive, but he is still attentive to his work.
In the old definition of this profile the contributor could be imprecise. But after having made a crowdsourcing campaign with real experts of the concerned domain, we noticed that in fact the experts are characterized by the precision and the speed of their answers.

\paragraph{The good contributor} He has less knowledge than the expert, which may lead him to doubt the answer to be given. 
In case of hesitation, the good contributor is imprecise in his contribution in order to be sure of it. 
He takes the necessary time to think about his answer and is attentive to his work.

\paragraph{The average contributor} He also has more limited knowledge than the expert about the task domain. This sometimes leads him, like the good contributor, to hesitate about which answer to select and thus to be imprecise. He is reflective and careful in his work.
However, unlike the good contributor, this imprecision does little to reinforce his certainty in his selection.

\paragraph{The bad contributor} This individual is not necessarily unqualified for the task, but he loses interest and responds as quickly as possible in order to finish the campaign as soon as possible.
He is one of the few contributors with a malicious profile who is only attracted by the greed and is not conscientious in his work.
Their answers are always precise, since this allows them to avoid wasting time. 
They are certain, because they want to compensate for their lack of seriousness with the employer by their self-confidence or because they do not have a good perception of their real abilities.
Nevertheless, it is possible to differentiate him from the expert thanks to his behavior as he is not attentive because his quick contributions are often random. This makes it more difficult for him to remember his answers during attention questions.

Once the mass functions $m_{c}^{\Omega_\mathcal{P}}$, $m_{c}^{\Omega_\mathcal{C}}$, $m_{c}^{\Omega_\mathcal{R}}$, $m_{c}^{\Omega_\mathcal{A}}$ have been calculated, a conversion of the frame of discernment $\Omega_\mathcal{P}$, $\Omega_\mathcal{C}$, $\Omega_\mathcal{R}$, $\Omega_\mathcal{A}$ is performed to get back to the frame of discernment on the profile $\Omega_\mathcal{XP}$.
Table \ref{tab:conversion} summarizes how the elements of the different frames of discernment are converted.

\begin{table}[t]
    \centering
    \begin{tabular}{|c|c|c|c|c|}
        \hline
        \bf{Characteristic} & \bf{$\Omega$} & \bf{Element} & \bf{Conversion}\\
        \hline
        \multirow{2}{*}{Imprecision} & \multirow{2}{*}{$\Omega_\mathcal{P}$} & $\mathcal{P}$ & $\{Expert, Bad \}$ \\
        \cline{3-4}
                    &            & $\mathcal{IP}$ & $\{Good, Average \}$ \\
        \hline
        \multirow{2}{*}{Certainty} & \multirow{2}{*}{$\Omega_\mathcal{C}$} & $\mathcal{C}$ & $\{Expert, Good, Bad \}$ \\
        \cline{3-4}
                  &            & $\mathcal{UC}$ & $\{Average \}$ \\
        \hline
        \multirow{2}{*}{Reflection} & \multirow{2}{*}{$\Omega_\mathcal{R}$} & $\mathcal{R}$ &  $\{Good, Average\}$ \\
        \cline{3-4}
                  &            & $\mathcal{NR}$ & $\{ Expert, Bad\}$ \\
        \hline
        \multirow{2}{*}{Attention} & \multirow{2}{*}{$\Omega_\mathcal{A}$} & $\mathcal{A}$ & $\{Expert, Good, Average \}$ \\
        \cline{3-4}
                  &            & $\mathcal{A}$ & $\{Bad\}$ \\
        \hline
    \end{tabular}
    \caption{Conversion of the frames of discernment $\Omega_\mathcal{P}$, $\Omega_\mathcal{C}$, $\Omega_\mathcal{R}$, $\Omega_\mathcal{A}$}
    \label{tab:conversion}
\end{table}

Thus, a person who is estimated precise $\mathcal{P}$ by MONITOR may be an $Expert$ or a $Bad$ contributor. On the contrary, if it is imprecise $\mathcal{IP}$ it is a $Good$ or $Average$ contributor.
Thanks to the conversion of the frame of discernment, it is then possible to combine the mass functions of qualification and behavior on $\Omega_\mathcal{XP}$. The combination yields a single mass function $m_c^{\Omega_\mathcal{XP}}$ for $c$:
\begin{equation}
    m_c^{\Omega_\mathcal{XP}} = \frac{\alpha_\mathcal{P} m_c^{\Omega_\mathcal{P} \rightarrow \Omega_\mathcal{XP}} 
    + \alpha_\mathcal{C} m_c^{\Omega_\mathcal{C} \rightarrow \Omega_\mathcal{XP}} 
    + \alpha_\mathcal{R} m_c^{\Omega_\mathcal{R} \rightarrow \Omega_\mathcal{XP}} 
    + \alpha_\mathcal{A} m_c^{\Omega_\mathcal{A} \rightarrow \Omega_\mathcal{XP}}}{\alpha_\mathcal{P}+\alpha_\mathcal{C}+\alpha_\mathcal{R}+\alpha_\mathcal{A}}
    \label{eq:profil}
\end{equation}
In equation \eqref{eq:profil}, $\Omega_i \rightarrow \Omega_\mathcal{XP}$ symbolizes the conversion of the frame of discernement $\Omega_i$ sur $\Omega_\mathcal{XP}$ provided by Table~\ref{tab:conversion}. 
The $\alpha_i$ coefficients are used to modulate the weight given to each element that defines the contributor profile. 
This function is then transformed into a pignistic probability in order to make a decision on the contributor's profile. \\

After the contributor's profile estimation, his answers are weakened according to his profile when aggregating the collected data in MONITOR phase 2.

\subsection{MONITOR phase 2: Answer Aggregation}

Thanks to the estimation of the contributor's profile, his answers are processed in order to be aggregated in the phase 2 schematized on figure~\ref{fig:MONITOR}.
The answer of the contributor $c$ to the question $q$ is modeled by the single support mass function $m_{cq}^{\Omega_q}$ given by equation \eqref{eq:X^w}. 
This function is then weakened by a value $\alpha_\mathcal{XP} \in [0,1]$ according to the contributor profil. 
The objective of this approach is to give more weight to the answers of qualified and conscientious contributors.
Thus, for an expert $\alpha_\mathcal{XP}$ will be close to 1.
In contrast, a bad contributor will receive a value $\alpha_\mathcal{XP}$ close to 0, so that its contributions do not negatively impact the quality of the data after aggregation.
Weakened mass functions $m_{cq}^{\Omega_q, \alpha_\mathcal{XP}}$ are combined for each question $q$ for the whole crowd: $m_{q}^{\Omega_q}$.
The functions $m_{q}^{\Omega_q}$ are finally transformed into pignistic probabilities in order to make a decision on the answer.

\section{Crowdsourcing campaigns}
\label{sec:campaigns}

In order to test MONITOR on real data we use four crowdsourcing campaigns performed by \cite{thierry22} described in the following paragraphs.
The main objective for these campaigns is always the same, a picture of a bird is presented to the contributor with a set of species names and he has to select the right answer.
As the users of the platform live in France, the birds used for the campaign are all of species visible in metropolitan France.
The Wirk platform (Crowdpanel\footnote{https://crowdpanel.io/ (15/04/2022)}) is used to realize the crowdsourcing campaigns.

\begin{table}[h]
    \centering
    \begin{tabular}{|l|c|c|c|}
        \hline
        Campaign & Answers & Crowd size & Number of answer \\
        \hline
        multi\_birds\_precise & Precise & 100 & 5000 \\
        multi\_birds\_imprecise & Imprecise & 100 &  5000 \\
        10\_birds\_precise & Precise & 50 & 2500\\
        10\_birds\_imprecise & Imprecise & 50 & 2500\\
        10\_birds\_dynamic & Imprecise & 51 & 2990\\
        \hline
    \end{tabular}
    \caption{Summary of crowdsourcing campaigns conducted}
    \label{tab:data_oiseaux}
\end{table}

\paragraph{Campaigns multi\_birds}
For these two campaigns, 5 bird names are proposed to the contributor, the names change from one question to another and a bird species is presented only once.
We have tried to introduce different levels of difficulty in the questions.
For example, for a difficult question, a photo of an eagle is presented to the contributor and the five answer items are different species of eagles.
Conversely, for a simpler question, a photo of a gull is presented to the contributor and the four other answers are names of duck species.
For a single photo, responses were presented in random order to each contributor to avoid selection bias. 
In addition, the questions were also asked in a random order, so that when a contributor $c_1$ answers a question $q_i$, $c_2$ answers $q_j$.

These crowdsourcing campaigns include 3 attention questions for which the contributor is asked to give the same answer as the one given in the previous question.
In both campaigns, the contributor has to give his answer, validate it, then specify his certainty according to the following Likert scale:
``Totally uncertain'',``Uncertain'', ``Rather uncertain'', ``Neutral'', ``Rather certain'', ``Certain'', ``Totally certain''.
Finally, after having given his answer and his certainty, he can validate his contribution in order to move on to the next question.
For the first campaign (multi\_birds\_precise) the contributor must give a precise answer by selecting a single bird name.
For the second campaign (multi\_birds\_imprecise) the contributor can be imprecise and select up to all of the bird names offered.
The crowds that contributed to both campaigns are composed of 100 contributors, and a contributor allowed to do the first campaign cannot participate in the second.
Each contributor must annotate 50 photos, for a total of 5000 contributions for each campaign.

These data are very useful for testing MONITOR, however, the 50 bird species to be identified are all distinct.
Moreover, among the four additional answers proposed to a question, some bird names are not part of the pictures presented in the other questions.
Thus, it is not possible to apply the EM algorithm. For this reason, additional crowdsourcing campaigns were conducted to collect the necessary data for a comparison of MONITOR with EM.

\paragraph{Campaigns 10\_birds}
For these campaigns, ten bird species are selected and proposed as response elements to the contributors.
In order to observe the contributor's ability to be imprecise in case of hesitation, the ten birds presented are composed of subgroups from the same bird family.
These ten names are presented to each contributor in a different order to avoid a possible selection bias.
This ordering of names is nevertheless fixed for a contributor throughout the campaign.
Such as the campaigns multi\_birds\_precise and multi\_birds\_imprecise the  questions are asked in a random order. The same scale is used for certainty and 3 attention questions are also asked.
The contributor is no longer required to validate his answer before he can give his certainty.

The crowdsourcing campaigns were conducted on the Crowdpanel platform and each campaign required the participation of 50 contributors.
The crowd size is halved because, as shown in the tests performed in the rest of this paper, a crowd of 50 contributors is sufficient to obtain a high correct response rate after data aggregation.
As with the other campaigns, a contributor who has participated in one experiment cannot participate in another.
For each of the ten bird species that make up the proposed answer set, the contributor is presented with 5 photos of a bird, so that the contributor answers 50 questions.
Thus 2500 data are collected for the experiments 10\_birds\_precise, with precise answers, and 10\_birds\_imprecise, for which the contributor can choose up to five answers.

There is 2990 data collected for 10\_birds\_dynamic because for this experiment, the contributor can give a second answer for the same question by modifying his first answer.
Indeed, if the contributor is imprecise in his answer, he is asked in a second step if he is able to restrict his choice of answer while giving his new certainty. 
When he is offered to restrict his selection, only the previously chosen answer elements are proposed again.
On the other hand, if he is precise but not ``totally certain'' of his answer, he is offered to widen his selection if he feels the need.
In this case, the first selected answer is kept in step 2 and he can complete it by selecting new names.
These interactions with the contributor increased the number of responses collected, and therefore the time spent soliciting the contributor. 

After introducing the data used to test MONITOR we present the experiments conducted to validate the model in the next section.

\section{Results}
\label{sec:resultats}

This section presents the results of our experiments.
First, we compare the precision and reflection estimation with state of the art methods.
Then in a second step we compare the profile estimation by MONITOR with that of \cite{rjab16} and EM.
Finally we compare the aggregation method of MONITOR with the MV and EM.

\subsection{Comparison of MONITOR estimations with existing}

MONITOR analyzes four elements to estimate the contributor's profile: imprecision, certainty, attention and reflection.
In order to validate our model we wanted to compare these four items with existing equivalent estimates.
However, to our knowledge, there is no method for estimating the certainty of the contributor for his response nor his attention during the campaign. We could not compare the proposed model on these two elements.
This section therefore presents comparisons of the calculation of contributor precision and reflection with existing methods.

\subsubsection{Comparison of $m_c^{\Omega_\mathcal{P}}(\mathcal{P})$ from MONITOR and $DP_c$ from \cite{rjab16}}

In order to calculate the mass function associated with the imprecision of the contributor $m_c^{\Omega_\mathcal{P}}$, given by equation \eqref{eq:mI}, we are inspired by the degree of precision $DP_c$ from \cite{rjab16}, equation \eqref{eq:DPc}.
Therefore, this section makes a comparison of the mass calculated by MONITOR to characterize the precision of the contributor $m_c^{\Omega_\mathcal{P}}(\mathcal{P})$ with the precision degree $DP_c$ of \cite{rjab16}.
As the answers $X \in 2^{\Omega_q}$ are modeled by simple supported mass functions, $DP_c$ can be written for $X$ different of $\Omega_q$ in this way:
\begin{equation}
    \displaystyle DP_c = \frac{1}{|E_{Q_c}|} \sum_{q \in E_{Q_c}} m_{cq}^{\Omega_q}(X) \left(1 - \frac{log_2(X)}{log_2(\Omega_q)} \right)
\end{equation}
\cite{thierry19} have $m_c^{\Omega_P}(P) = \alpha*DP_c$ and $m_c^{\Omega_P}(IP) = \alpha*(1 - DP_c)$ with $\alpha$ a discounting coefficient. 
So comparing $m_c^{\Omega_P}(P)$ with $DP_c$ is equivalent to compare it with the old calculation of \cite{thierry19}.

Since $m_c^{\Omega_\mathcal{P}}(\mathcal{P})$ and $DP_c$ calculate the imprecision of the contributor, the comparison is made on the data collected during the campaigns where the contributor can be imprecise: multi\_birds\_imprecise, 10\_birds\_imprecise, 10\_birds\_dynamic.
For the campaign multi\_birds\_imprecise, $|\Omega_q|=5$ since the contributor is offered five answers to each question, while for 10\_birds\_imprecise and 10\_birds\_dynamic, $|\Omega_q|=10$ as 10 answers are offered.
For these three campaigns the contributor can choose to select up to 5 bird names, so $imp_{MAX}=5$ in each case. 

\begin{figure}
    \centering
    \includegraphics[height=8cm]{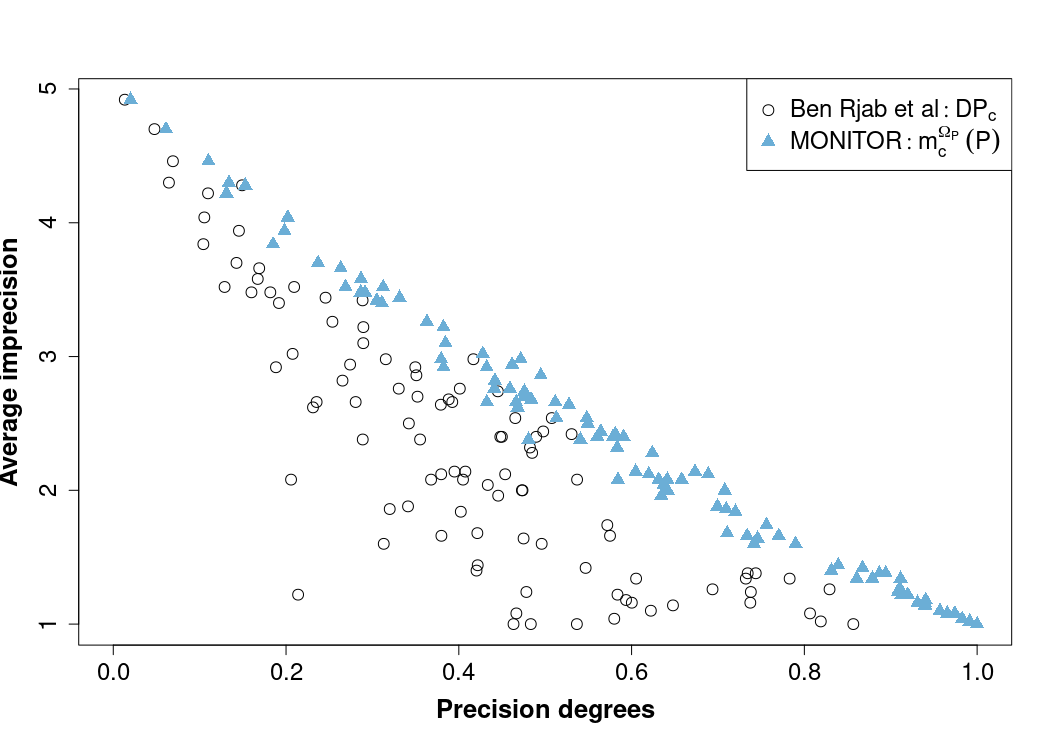}
    \caption{Degree comparison for the campaign {\bf multi\_birds\_imprecise}.}
    \label{fig:DP_vs_mP}
\end{figure}

\begin{figure}
    \centering
    \includegraphics[height=8cm]{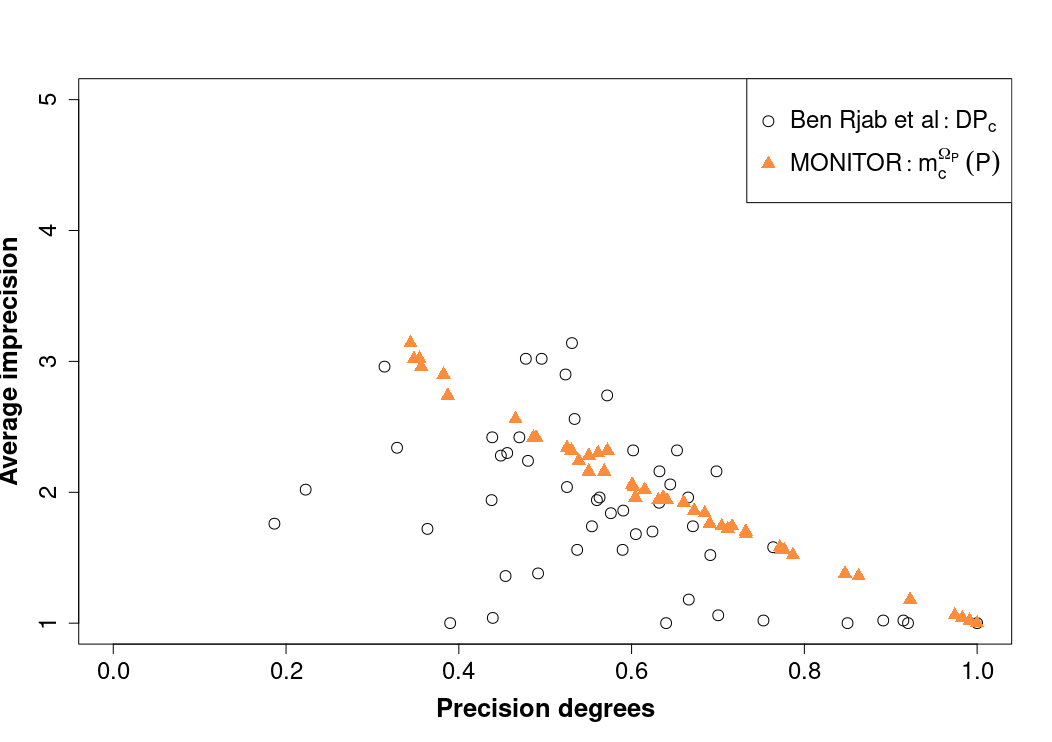}
    \caption{Degree comparison for the campaign  {\bf 10\_birds\_imprecise}.}
    \label{fig:DP_vs_mP_10_w2}
\end{figure}

\begin{figure}
    \centering
    \includegraphics[height=8cm]{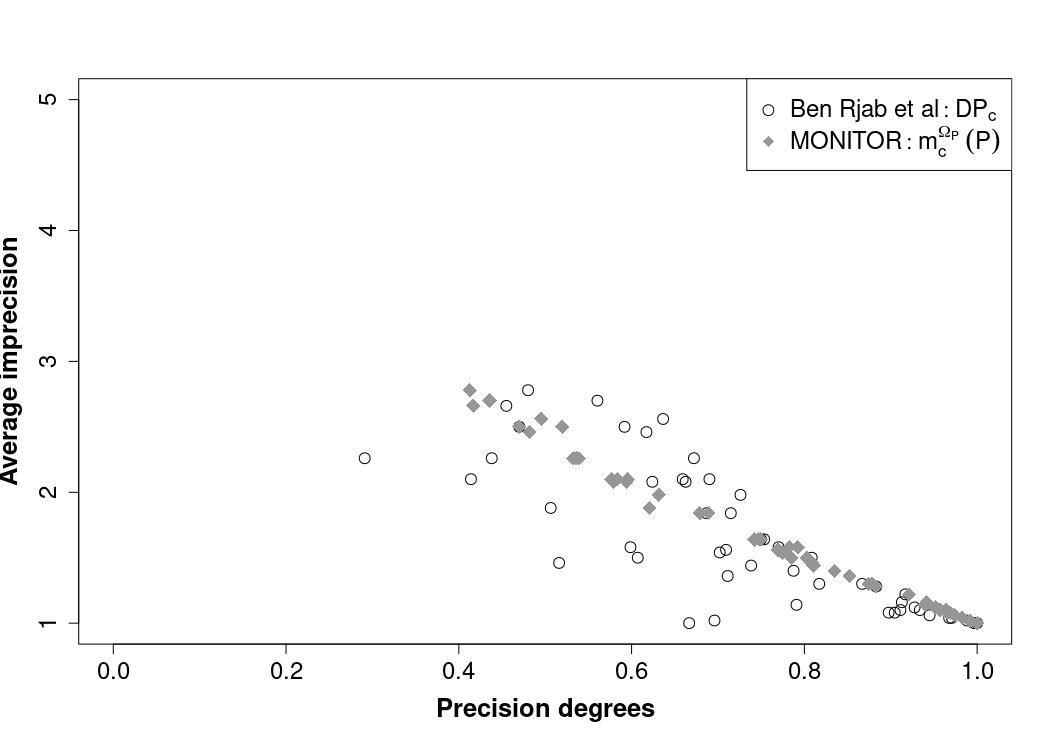}
    \caption{Degree comparison for the campaign  {\bf 10\_birds\_dynamic}.}
    \label{fig:DP_vs_mP_10_w3}
\end{figure}

Figures \ref{fig:DP_vs_mP}, \ref{fig:DP_vs_mP_10_w2} and \ref{fig:DP_vs_mP_10_w3} present comparisons of $m_c^{\Omega_\mathcal{P}}(\mathcal{P})$ and $DP_c$ for the three campaigns.
For these three figures, a contributor $c$ is represented by two points of the same ordinate which corresponds to its average imprecision.
These points have as abscissa the value of $DP_c$ for one and that of $m_c^{\Omega_\mathcal{P}}(\mathcal{P})$ for the other.
For example, in Figure~\ref{fig:DP_vs_mP} the two points that have an average imprecision of 5 on the ordinate and values $DP_c$ and $m_c^{\Omega_\mathcal{P}}(\mathcal{P})$ both correspond to the same contributor.
As $imp_{MAX}=5$, the imprecision varies between 1 and 5, and the values of $DP_c$ and $m_c^{\Omega_\mathcal{P}}(\mathcal{P})$ are included in this interval $[0,1]$.

On figure~\ref{fig:DP_vs_mP}, $DP_c \leq m_{c}^{\Omega_\mathcal{P}}(\mathcal{P})$ because $imp_{MAX}=|\Omega_q|$. 
We see in this figure that the more precise a contributor is on average, the closer its imprecision is to 1 on the graph, the larger the gap between the values of $DP_c$ and $m_{cq}^{\Omega_q}$ is growing.
For example, a contributor has a degree $DP_c$ close to 0.2 on this graph while its average imprecision is very close to 1 as well as $m_{c}^{\Omega_\mathcal{P}}(\mathcal{P})$. 
As for this graph $imp_{MAX}=|\Omega_q|$, the gap between the values of $DP_c$ and $ m_{c}^{\Omega_\mathcal{P}}(\mathcal{P})$ is only due to the mass $m(X)$ in the computation of $DP_c$ which reduces the quality of the estimate of the contributor's imprecision.

On figures \ref{fig:DP_vs_mP_10_w2} and \ref{fig:DP_vs_mP_10_w3},  $imp_{MAX} < |\Omega_q|$ that is why $m_{c}^{\Omega_\mathcal{P}(\mathcal{P})} \leq DP_c$ in the case where $\gamma(|X|) \leq m_{cq}^{\Omega_q}(X)$ with:
\begin{equation}
    \displaystyle \gamma(|X|) = \frac{log_2|\Omega_q|}{log_2(imp_{MAX})} \frac{log_2(imp_{MAX}) - log_2|X|}{log_2|\Omega_q| - log_2|X|}
\end{equation}
Table \ref{tab:val_gamma} lists the numerical values obtained for $\gamma$ for $imp_{MAX}=5$ and $|\Omega_q|=10$.
\begin{table}[t]
    \centering
    \begin{tabular}{|c|c|c|c|c|c|c|}
        \hline
        $|X|$ & 1 & 2 & 3 & 4 & 5   \\
        \hline
        $\gamma(|X|)$ & 1.00 & 0.82 & 0.61 & 0.35 & 0.00 \\
        \hline
    \end{tabular}
    \caption{Value of $\gamma(|X|)$ based on the imprecision of the answer.}
    \label{tab:val_gamma}
\end{table}
The condition $\gamma(|X|) \leq m_{cq}^{\Omega_q}(X)$ is for example verified when $|X|=4$ and the contributor is certain of his answer, which corresponds to $m_{cq}^{\Omega_q}(X)=0.83$.

We observe in figures \ref{fig:DP_vs_mP_10_w2} and \ref{fig:DP_vs_mP_10_w3} that the average imprecision does not exceed 3 while it reaches a value close to 5 for some contributors of figure~\ref{fig:DP_vs_mP}.
This is due to the fact that although it is possible for the contributor to select up to 5 birds for the campaigns 10\_birds\_imprecise and 10\_birds\_dynamic this is not necessary.
Indeed, the 10 proposed species can be grouped according to their family, i.e. 1 Muscicapidae, 2 Columbidae, 3 Paridae and 4 Corvidae.
It is possible to hesitate between birds of the same family but more difficult with birds of different families, the pigeon and the crow are for example very distant.
This is why the average imprecision should not exceed 4 in logic, which is consistent with the observed results.

On figures~\ref{fig:DP_vs_mP_10_w2} and \ref{fig:DP_vs_mP_10_w3}, the values of $DP_c$ are getting closer to $m_c^{\Omega_\mathcal{P}}(\mathcal{P})$ in comparison with the figure \ref{fig:DP_vs_mP}, especially for figure~\ref{fig:DP_vs_mP_10_w3} whose data is provided by the dynamic campaign.
However, the values of the two imprecision of the estimates remain distinct because of the use of $m(X)$ in the computation of $DP_c$, but also because $imp_{MAX} < |\Omega_q|$.
Overall, according to the graphs,  $m_c^{\Omega_\mathcal{P}}(\mathcal{P})$ is more representative of the average of the contributor's imprecision than $DP_c$. 

In this section we performed a comparison of the estimation of a contributor's imprecision with the state of the art. 
In particular, we have seen that the estimation of imprecision by MONITOR is more relevant than the calculation of \cite{rjab16} and therefore of \cite{thierry19} as well.
In the next section we do the same for the contributor's reflection for the task.

\subsubsection{Comparison of reflection estimation with a statistical approach}

MONITOR uses the contributor's response time to a question $T_{cq}$ and compares it to an expected minimum time $T_{0q}$ to compute the mass function associated with the contributor's reflection.

We compare MONITOR's reflection calculation with the statistical method of excluding marginal contributors of \cite{komarov13} because they also use the response time.
According to the authors, a contributor is considered marginal if his response time is too far from the response time of the whole crowd.
A contributor who responds too quickly is suspected of responding randomly and a contributor who is too long is deemed irrelevant by the authors.
To determine the marginal contributors, \cite{komarov13} calculate the interquartile range (IQR) of response times, which is the difference between the third ($Q3$) and the first quartile ($Q1$).
A contributor is considered marginal and therefore excluded from the authors' study if his response time calculation is not included in the interval $[Q1 - 3*IQR, Q3 + 3*IQR]$.
%
A contributor's marginality rate is calculated basing on the exclusion method of \cite{komarov13} in order to obtain an element of comparison with the reflection.
To calculate this validity rate, for each question, an indicator function is associated with the validity of the contributor's answer.
If $T_{cq} \in [Q1 - 3*IQR, Q3 + 3*IQR]$ the contribution is validated and the indicator function is 1.
The average of the indicator's functions is performed for each contributor to obtain the average validity.

In the experiments conducted with MONITOR, $T_{0q}$ is equal to the first quartile of all the response times of the crowd to the question $q$, and $\alpha_R=0.9$ arbitrarily.
For each contributor the average of the mass functions $m_{cq}^{\Omega_\mathcal{R}}$ is calculated to obtain $m_{c}^{\Omega_\mathcal{R}}$ and the pignistic probability $Betp(\mathcal{R})$ is calculated.
The probability $BetP(\mathcal{R})$ that the contributor is reflective is then compared to the contributor's average validity rate.

\begin{figure}[t]
    \centering
     \begin{subfigure}[b]{0.48\textwidth}
         \centering
    \includegraphics[width=\textwidth]{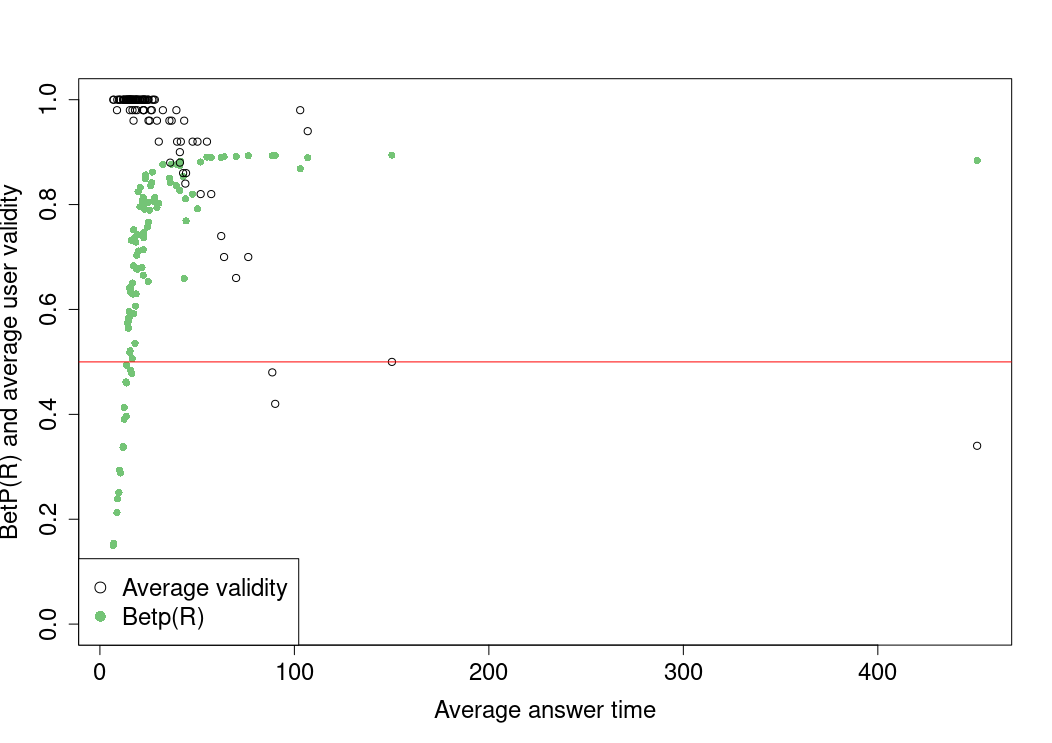}
    \caption{multi\_birds\_precise}
    \label{fig:BetPvsVm_xp1}
     \end{subfigure}
    \quad
     \begin{subfigure}[b]{0.48\textwidth}
         \centering
    \includegraphics[width=\textwidth]{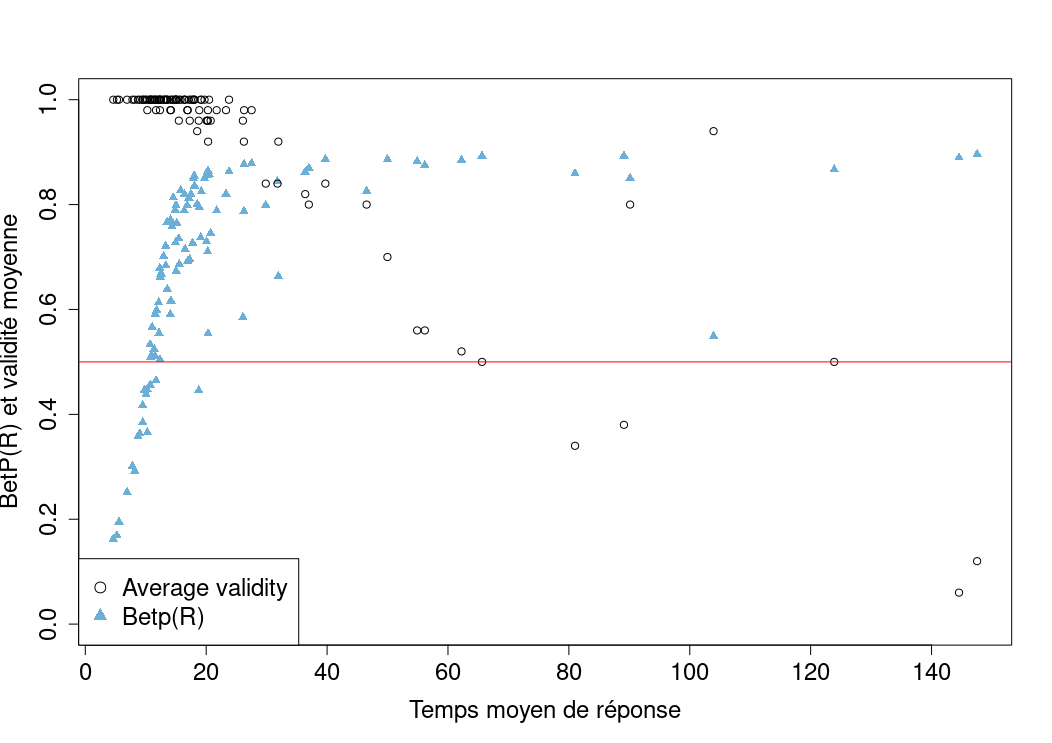}
    \caption{multi\_birds\_imprecise}
    \label{fig:BetPvsVm_xp2}
     \end{subfigure}
    \caption{Pignistic probability that the contributor is reflective and average contributor validity for experiments multi\_birds.}
    \label{fig:BetPvsVm_multi}
\end{figure}

\begin{figure}[t]
    \centering
     \begin{subfigure}[b]{0.48\textwidth}
         \centering
    \includegraphics[width=\textwidth]{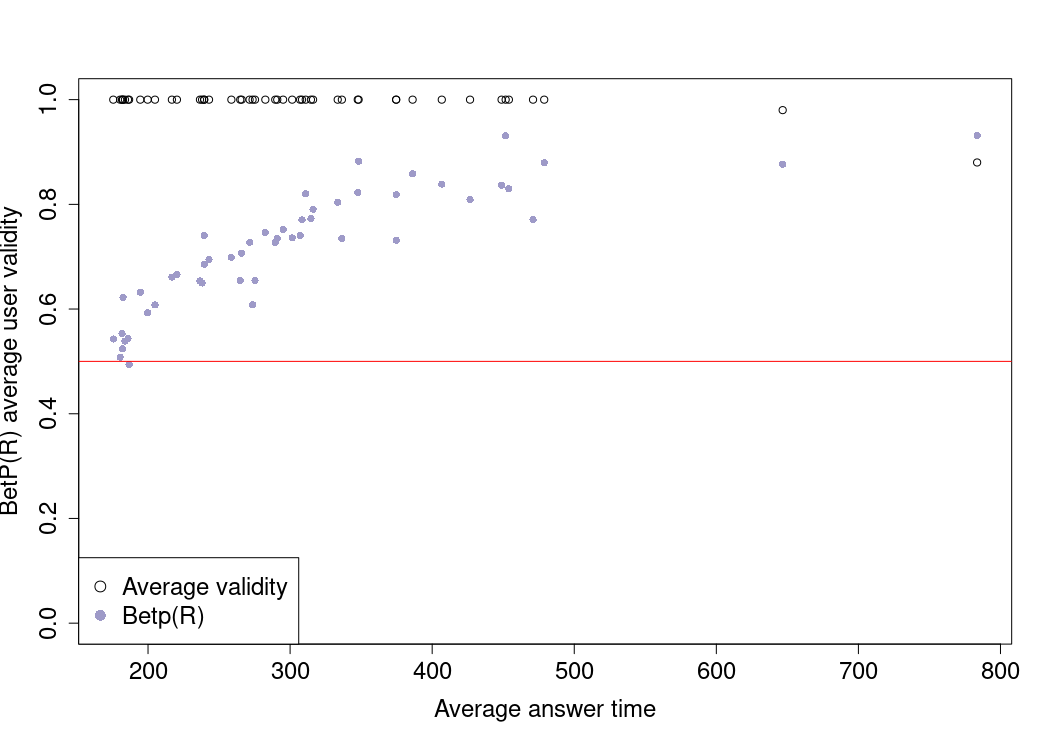}
    \caption{10\_birds\_précis}
    \label{fig:BetPvsVm_xp1_10}
     \end{subfigure}
    \quad
     \begin{subfigure}[b]{0.48\textwidth}
         \centering
    \includegraphics[width=\textwidth]{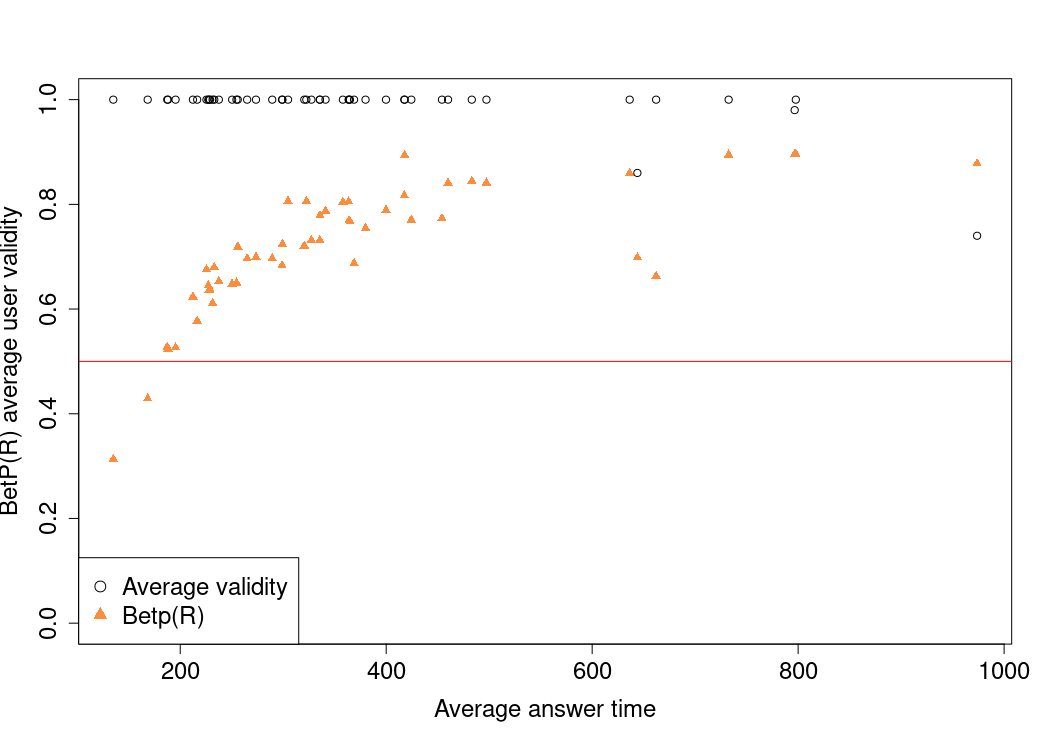}
    \caption{10\_birds\_imprecise}
    \label{fig:BetPvsVm_xp2_10}
    \end{subfigure}
     \begin{subfigure}[b]{0.48\textwidth}
         \centering
    \includegraphics[width=\textwidth]{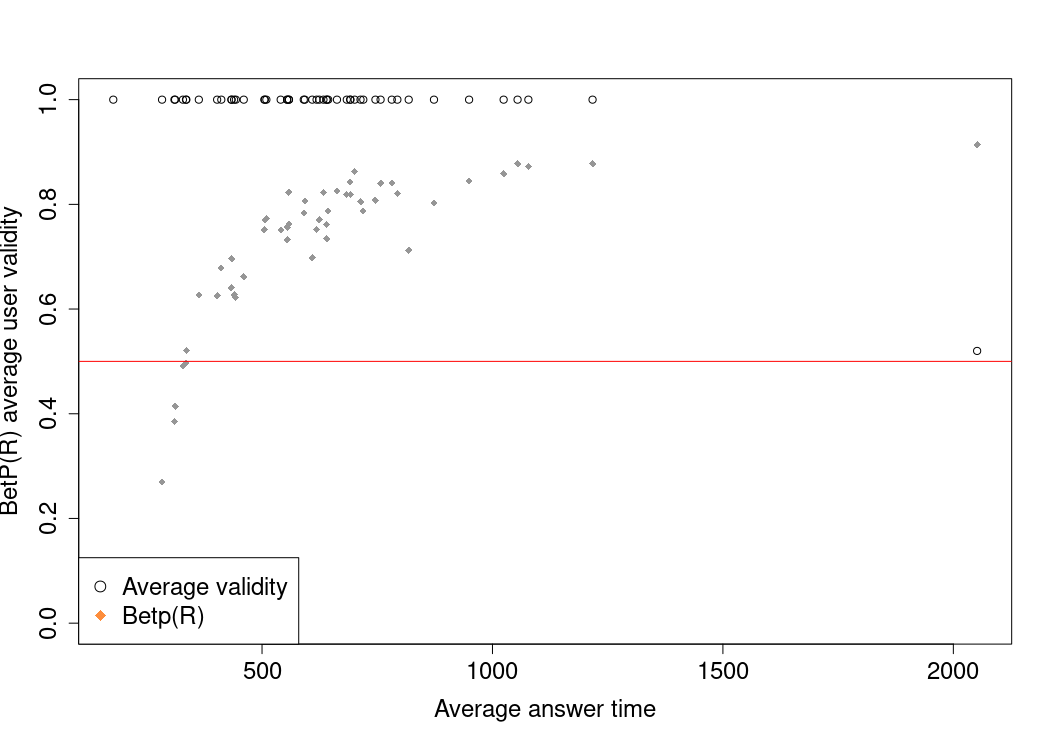}
    \caption{10\_birds\_imprecise}
    \label{fig:BetPvsVm_xp3_10}
    \end{subfigure}
    \caption{Pignistic probability that the contributor is reflective and average contributor validity for experiments 10\_birds.}
    \label{fig:BetPvsVm_10}
\end{figure}

For each contributor, the average time to answer questions $T_c$ is computed for the whole campaign. 
This is the average time presented by the x-axis for figures \ref{fig:BetPvsVm_multi} and \ref{fig:BetPvsVm_10}.
The y-axis corresponds for a contributor respectively to the pignistic probability that is reflected during the whole campaign and to the average validity rate inspired by the selection method of \cite{komarov13}.
On each graph in these figures a contributor is represented by two points, one point for the estimated reflection and another for its average validity.
On each graph in these figures a contributor is represented by two points, one point for the estimated reflection and another for its average validity.
For this two experiments shown figure~\ref{fig:BetPvsVm_multi}, the majority of contributors have short average response times which facilitates the appearance of marginal contributors with longer average response times.
This is contrary to the experiments with 10 birds in figure~\ref{fig:BetPvsVm_10} where the average response times are more dispersed.

According to figures~\ref{fig:BetPvsVm_multi} and \ref{fig:BetPvsVm_10} some contributors are considered not reflected during the campaign by MONITOR, $BetP(R)<0.5$.
This means that either they did not take the time to think because they respond randomly, or they have above-average expertise on the task and need less time to think.

\begin{table}
    \centering
    \begin{tabular}{|l|c|c|c|c|c|}
        \hline
         &  \multicolumn{2}{c|}{Average validity $<$ 0.5} & \multicolumn{2}{c|}{Average validity $\geq$ 0.5} & \\
         \cline{2-5}
         Campaigns & {\footnotesize GRRec $<$ 0.5}  &  {\footnotesize GRRec $\geq$ 0.5} & {\footnotesize GRRec$<$ 0.5} &  {\footnotesize GRRec$\geq$ 0.5}  & CCR \\
         \hline
         {\footnotesize multi\_birds\_precise} & 0 & 4 & 66 & 30 & 30 \\
         \hline
         {\footnotesize multi\_birds\_imprecise} & 1 & 2 & 71 & 27 & 28 \\
         \hline
         {\footnotesize 10\_birds\_precise} & 0 & 0 & 38 & 12 & 24 \\
         \hline
         {\footnotesize 10\_birds\_imprecise} & 0 & 0 & 6 & 44 & 88 \\
         \hline
         {\footnotesize 10\_birds\_dynamic} & 0 & 0 & 4 & 47 & 92 \\
         \hline
    \end{tabular}
    \caption{Comparison of contributors' average validity and their correct recognition rates.}
    \label{tab:matConf_tauxVal}
\end{table}

\begin{table}
    \centering
    \begin{tabular}{|l|c|c|c|c|}
        \hline
         &  \multicolumn{2}{c|}{betP($\mathcal{R}$)$<$ 0.5} & \multicolumn{2}{c|}{betP($\mathcal{R}$) $\geq$ 0.5}\\
         \cline{2-5}
         Campaigns & {\footnotesize GRRec $<$ 0.5} & {\footnotesize GRRec $\geq$ 0.5} & {\footnotesize GRRec$<$ 0.5} &  {\footnotesize GRRec$\geq$ 0.5} \\
         \hline
         multi\_birds\_precise & 0 & 0  & 66 & 34 \\
         \hline
         multi\_birds\_imprecise & 0 & 0 & 72 & 29\\
         \hline
         10\_birds\_precise & 0 & 0 & 38 & 12 \\
         \hline
         10\_birds\_imprecise 0 & 0 & 0 & 6 & 44 \\
         \hline
         10\_birds\_dynamic &  0 & 0 & 4 & 47\\
         \hline
    \end{tabular}
    \caption{Comparison of the old pignistic probabilities of \cite{thierry19} that contributors are reflective with their rate of good bird recognition.}
    \label{tab:matConf_BetPR_ICTAI}
\end{table}

\begin{table}
    \centering
    \begin{tabular}{|l|c|c|c|c|}
        \hline
         &  \multicolumn{2}{c|}{betP($\mathcal{R}$)$<$ 0.5} & \multicolumn{2}{c|}{betP($\mathcal{R}$) $\geq$ 0.5}\\
         \cline{2-5}
         Campaigns & {\footnotesize GRRec $<$ 0.5} & {\footnotesize GRRec $\geq$ 0.5} & {\footnotesize GRRec$<$ 0.5} &  {\footnotesize GRRec$\geq$ 0.5} \\
         \hline
         multi\_birds\_precise & 17 & 0 & 49 & 34 \\
         \hline
         multi\_birds\_imprecise & 17 & 0 & 55 & 29\\
         \hline
         10\_birds\_precise & 1 & 0 & 37 & 12 \\
         \hline
         10\_birds\_imprecise & 0 & 2 & 6 & 42 \\
         \hline
         10\_birds\_dynamic & 2 & 4 & 2 & 43 \\
         \hline
    \end{tabular}
    \caption{Comparison of the pignistic probabilities that contributors are reflective with their rate of good bird recognition.}
    \label{tab:matConf_BetPR}
\end{table}

In tables~\ref{tab:matConf_tauxVal}, \ref{tab:matConf_BetPR_ICTAI} and \ref{tab:matConf_BetPR}, we make a comparison, of the average validity for the first one and of the pignistic probability on the reflection for the second ones, with the good recognition rate of the contributor (GRRec).
For the average validity, \cite{komarov13} exclude from their analysis the contributors for which the response time is not included in the desired time interval.
Thus, we calculate the correct classification rate (CCR) between the average validity and the GRRec of the contributors.
For reflection, it is not possible to calculate this percentage of correct classification immediately because among the non-reflective contributors there are both malicious contributors and experts.
From Table \ref{tab:matConf_BetPR_ICTAI}, the reflection estimate based on the method employed by \cite{thierry19} does not identify any contributors with not-reflective responses. However, with the new MONITOR approach, not only are contributors estimated to be not-reflective, but their correct response rate is less than 0.5. Thus, there has been an improvement in the estimation of reflection since \cite{thierry19}.

As we have observed in figures~\ref{fig:BetPvsVm_multi} and \ref{fig:BetPvsVm_10}, all contributors of the experiments with 10 birds have an average validity greater than or equal to 0.5.
According to \ref{tab:matConf_tauxVal}, the best percentages of correct classification are obtained for the campaigns 10\_birds\_imprecise and 10\_birds\_dynamic for which the same 10 bird's names are always proposed to the contributor and it can be imprecise.
For the campaign multi\_birds\_imprecise, the contributor can also be imprecise but the proposed bird's names change with each question, but for this campaign 72 contributors have a GRRec$<0.5$.
In addition to allowing the contributor to be imprecise, using the same answer suggestions for all questions seems to improve the GRRec.
This finding is the same for the GRRecs in Table~\ref{tab:matConf_BetPR}.

For the campaigns multi\_birds, precise and imprecise, contributors estimated not to be reflective by MONITOR have a GGRec lower than 0.5 and are therefore a priori marginal contributors.
However, these marginal contributors have a validity rate higher than 0.5 according to table \ref{tab:matConf_tauxVal} when they should also be estimated as marginal.
The problem of the statistical method of \cite{komarov13} is that it can identify contributors as marginal if their response times are far enough from $[Q1-3*ICQ, Q3+3*ICQ]$.
This is visible in figure~\ref{fig:BetPvsVm_multi}, where the average response times are very short and the validity rates are grouped.
Rates below 0.5 represent contributors who have a much higher average response time than others. 

The methode of \cite{komarov13} does manage to identify marginal contributors here, but mainly those with a high average response time and a GRRec greater than or equal to 0.5.
These are therefore contributors whose answers are relevant, which is not beneficial to the aggregation of contributions. 
Conversely, some contributors are validated by this statistical approach but have a GRRec less than 0.5.
MONITOR identifies more non-reflective contributors, and this category includes both malicious and expert contributors, which is why some have $GRRec<0.5$ and others $GRRec\geq0.5$. 

After having made in this section a comparison of the imprecision and the reflection of the contributor with the existing one, we present in the following the experiments carried out on the profile in its entirety.

\subsection{Results for the profile estimation}

This section reviews the tests performed for the estimation of the contributor profile by MONITOR and the comparisons made with \cite{rjab16} and EM.
For these experiments only data including imprecise responses are used since they are more relevant for the profile as it is not interesting to calculate $m_c^{\Omega_\mathcal{P}}$ on precise data.
The data manipulated here comes from the campaigns: multi\_oiseau\_imprecise, 10\_birds\_imprecise and 10\_birds\_dynamic.

\paragraph{Semi-supervised learning to determine $\alpha_\mathcal{XP}$}
In order to calculate the mass on the contributor's profile $m_c^{\Omega_\mathcal{XP}}$, a conversion of the discernment frameworks associated with the qualification and behavior of the contributor is performed.
The converted mass functions are weighted by coefficients $\alpha_x$ as indicated by equation \eqref{eq:profil} page \pageref{eq:profil}.
It is not trivial to define the weights to be given to each element of the profile, which is why we split the results of our campaigns into two data sets in order to perform semi-supervised learning.
The pignistics probabilities on the contributors' profile are calculated for values $\alpha_\mathcal{P}$, $\alpha_\mathcal{C}$, $\alpha_\mathcal{R}$, $\alpha_\mathcal{A} \in [0,10]$. 
For each $X$ answer filled in to a $q$ question, the boolean $isValid$ that indicates the validity of the answer is divided by the number $|X|$ of items selected by the contributor.
The average of these quotients over all the questions allows us to obtain the contributor's correct response rate:
\begin{equation}
    CRR_c = \sum_{q \in E_{Q_c}} \frac{isValid}{|X|}
    \label{eq:CRRc}
\end{equation}
This $CRR_c$ rate is used to calculate the correct classification rates of MONITOR and the other approaches it is compared to.

\begin{table}[]
    \centering
    \begin{tabular}{|c|c|c|c|c|}
    \hline
                & Bad    & Average         & Good             & Expert  \\
    \hline
        $CRR_c$ & $\leq 0.2$ & $]0.2,0.5[$ & $[0.5,0.85 [$ & $\geq 0.85$ \\
    \hline
    \end{tabular}
    \caption{Summary table of profiles based on the values of the contributors' correct response rates ($CRR_c$).}
    \label{tab:profil_ref}
\end{table}

Before deploying the online crowdsourcing campaigns, three ornithologists conducted the multi\_birds\_imprecise in order to validate them.
The average correct response rate of the Experts during the campaign is 0.89, so we consider that a contributor whose $CRR_c \geq 0.85$ can be recognized as an expert in bird classification for this task.
Experts are not required to have a correct response rate of 1.00 because some species or photos may make it more difficult to identify the bird.
Arbitrarily, we consider that a person able to identify every second bird or more ($CRR_c \geq 0.5$) is a good contributor.
On the other hand, a contributor who identifies only one bird out of the five proposed or less ($CRR_c \leq 0.2$) is a Bad contributor because his answers are random.
Finally, an Average contributor has a correct answer rate included between the Good and the Bad contributor: his answers are less relevant than the Good contributor but not random.
All this information is summarized in table \ref{tab:profil_ref}.

\begin{table}[]
    \centering
    \begin{tabular}{|c|c|c|c|c|c|}
        \hline
         & Min & Q1 & Average & Q3 & Max  \\
        \hline
        multi\_birds\_imprecise & 0.16 & 0.30 & 0.44 & 0.52 & 0.96 \\
        \hline
        10\_birds\_imprecise & 0.29 & 0.39  & 0.44 & 0.46 & 0.92 \\
        \hline
        10\_birds\_dynamic & 0.36 & 0.45 & 0.52 & 0.55 & 0.92 \\
        \hline
    \end{tabular}
    \caption{Summary of contributors' correct response rates for the  {\bf training data}.}
    \label{tab:summary_apprentissage}
\end{table}

For the training data of the campaign multi\_birds\_imprecise, the minimum of the $CRR_c$ values is 0.16 according to table \ref{tab:summary_apprentissage} and the maximum of 0.96 with a third quartile of 0.52, which means that this training data set does include the four profile types to be identified.
This is not the case for training data of the campaigns 10\_birds, it seems that no Bad contributor participated.
This may be due to the smaller crowd size for these two campaigns.

\begin{table}[]
    \centering
\begin{tabular}{|c|c|c|c|c|c|c|}
    \hline
    \multirow{2}*{Data} & \multicolumn{4}{|c|}{Coefficients $\alpha$} & \multicolumn{2}{|c|}{CCR}\\
    \cline{2-7}
    & $\alpha_I$ & $\alpha_R$ & $\alpha_C$ & $\alpha_A$ & Learning & Test  \\
    \hline
    multi\_birds\_imprecise & 1 & 7 & 1 & 1 & 0.58 & 0.55 \\
    \hline
    10\_birds\_imprecise & 1 & 5 & 3 & 1 & 0.68 & 0.40 \\
    \hline
    10\_birds\_dynamics & 2 & 6 & 1 & 1 & 0.28 & 0.36 \\
    \hline
\end{tabular}
    \caption{Data Learning}
    \label{tab:apprentissage}
\end{table}

Table \ref{tab:apprentissage}, presents the values of coefficients $\alpha$ retained after training, and the correct classification rates ($CCR$) on the training and test data.
The values of $\alpha$ tested range from 0 to 10.
For the campaign multi\_birds\_imprecise $CCR=0.58$ for the training data and $CCR=0.55$ for the test data, which is encouraging given the complexity of estimating the contributor's profile in the absence of gold data.
The $CCR$ of the training data of 10\_birds\_imprecise is higher with 0.68\% of correct classification but the result on the test data is less good.
Finally, for the campaign 10\_birds\_dynamic, the $CCR$ are the worst, in the best cases we obtain $CCR=0.28$ for the training data.
For this campaign the contributor is sometimes asked if it is possible to refine or enlarge his selection of bird's names, the possible modified answer of the contributor is not considered here.
MONITOR does not seem to fit this campaign as well as others probably due to the iterative aspect of the questionnaire which is not currently considered by the model.

We note that the coefficient $\alpha_\mathcal{A}$ is low since it is equal to 1 for optimal classification rates during training, and it is the coefficient $\alpha_\mathcal{R}$ that is most important.
This means that the mass functions on reflection have more weight for profile estimation than those on attention.
Imprecision and certainty are of different importance depending on the data but are less important than reflection.

\paragraph{Comparison with the existing for the estimation of the expertise} 
\cite{rjab16} also use the theory of belief functions to determine the profile of the contributor, which is why we compare here the estimate made by MONITOR with that of the authors.
Another approach to profile estimation is to consider the confusion matrix on contributor's responses constructed by EM.
MONITOR considers in addition to the qualification of the contributor his behavior to estimate his profile, which is not the case for the two other methods. 

In order to estimate the profile of the contributor $c$, \cite{rjab16} calculate the degree of exactitude of the answers $DE_c$ and the degree of precision $DP_c$ given by equations \eqref{eq:DEc} and \eqref{eq:DPc} page \pageref{eq:DEc}.
The authors then propose to perform a clustering thanks to k-mean with $k=2$ to differentiate expert contributors from non-experts.
The article indicates that among the two sets obtained, the one with the higher average value of $DE_c$ is composed of experts. 
We use for the estimation of the profiles, according to the approach of \cite{rjab16}, the k-mean algorithm with $k=4$, as MONITOR defines four types of contributor profiles. 
Since for the authors the group of experts has the highest average value $DE_c$, we also consider an increasing expertise of the contributor's profile according to $DE_c$. 
The correct classification rates for the clustering on DE,DP and that on DG are included in table \ref{tab:profil_ref}.

It is possible to estimate the expertise of the contributors by using the confusion matrices calculated by the EM algorithm.
In the experiments, the positive predictive value (PPV) of the contributor is calculated thanks to its confusion matrix.
Once the PPV values obtained for the whole crowd, a clustering is performed with $k=4$.
The results of the clustering are then compared to the expected values in table \ref{tab:profil_ref}. 

\begin{table}[]
    \centering
    \begin{tabular}{|l|c|c|c|c|}
        \hline
        Campaign               & MONITOR & Clustering DE,DP & Clustering DG & EM  \\
        \hline
        multi\_birds\_imprecise &  0.43   &   0.17           &  0.4  & / \\
        10\_birds\_imprecise   &  0.54   &   0.24           &  0.42 & 0.62 \\
        10\_birds\_dynamic  &  0.31   &   0.14           &  0.51 & 0.61  \\
        \hline
    \end{tabular}
    \caption{Summary of the correct classification rates of the profile for: MONITOR, \cite{rjab16}, EM.}
    \label{tab:comp_profil}
\end{table}

It is not possible to apply EM to the data from the multi\_birds\_imprecise campaign because the answers proposed to question $q$ are not the same as those proposed to question $q+1$, which prevents the confusion matrices from being established.
Thus there is no value in the row associated with this campaign for EM. 
We note for the campaigns 10\_birds\_imprecise and 10\_birds\_dynamic that EM is the approach that offers the best results for profile estimation compared to MONITOR and \cite{rjab16}. 
The difference between the correct classification rates is 0.08 for the campaign 10\_birds\_imprecise but 0.3 for 10\_birds\_dynamic because the profile estimate by MONITOR is the worst for this campaign.\\

The use of EM to estimate the contributor's profile is more relevant than MONITOR, especially in the case of crowdsourcing campaigns where a question can be asked again to the contributor to evolve his answer.
In the case of a simpler campaign where the contributor is only asked once, the difference in correct classification rates between EM and MONITOR remains small.
Moreover, as explained in this section, EM unfortunately cannot be used if the set of answers proposed to the contributor changes from one question to another unlike MONITOR which is not impacted by this. 
The purpose of estimating the contributor's profile is to weaken his answers according to his profile, so the following section presents the experiments performed for the aggregation of contributions.

\subsection{Aggregation results}

First, this section presents the comparison of different combination operators applied to the mass functions modeling the contributors' responses $m_{cq}^{\Omega_q}$.
Then this method will be used by MONITOR for the aggregation of the contributors' responses and compared to MV and EM.\\

First of all, tests are carried out to determine the coefficient $\alpha$ used to weaken $m_{cq}^{\Omega_q}$.
To do this, the responses are combined by the normalized conjunctive operator, equation \eqref{eq:mConjNorm}, for the five data sets for values $\alpha \in [0.1,1]$.
The conjunctive operator is chosen because the discounting of responses has more impact for it, unlike the mean or the LNS rule.
Tests show that the best results are obtained for $\alpha=0.8$ and $\alpha=0.9$. 
We have chosen to use $\alpha=0.9$ for the discounting of the mass functions before their combination for the comparison of the operators.

\begin{figure}[t]
    \centering
    \includegraphics[height=8cm]{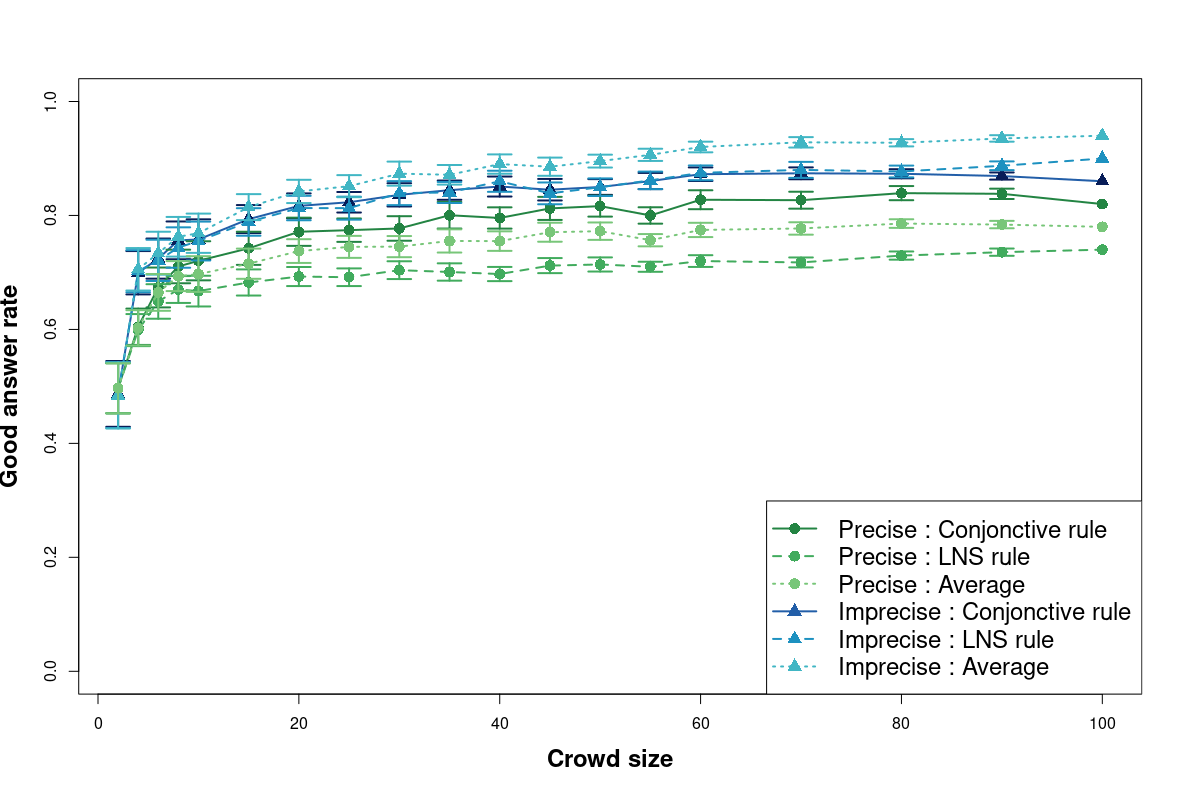}
    \caption{Comparison of operators: Conjunctive, LNS and mean for the campaigns {\bf multi\_birds\_precise and multi\_birds\_imprecise.}}
    \label{fig:CRR_Foule_Op_multi}
\end{figure}

On figure \ref{fig:CRR_Foule_Op_multi} the aggregation of the responses is performed for each operator and for an increasing size $n$ of the crowd, such that:
$$n \in \{2,4,6,8,10,15,20,25,30,35,40,45,50,55,60,70,80,90,100\}$$
For each value of $n$ the contributors whose responses are aggregated are randomly selected, then the CRR is computed on the 50 pictures.
This process of crowd selection, aggregation of contributions and calculation of CRRs is performed 50 times for each value of $n$ in order to obtain a mean CRR for $n$ and a 95\% confidence interval.
\begin{figure}[t]
    \centering
    \includegraphics[height=8cm]{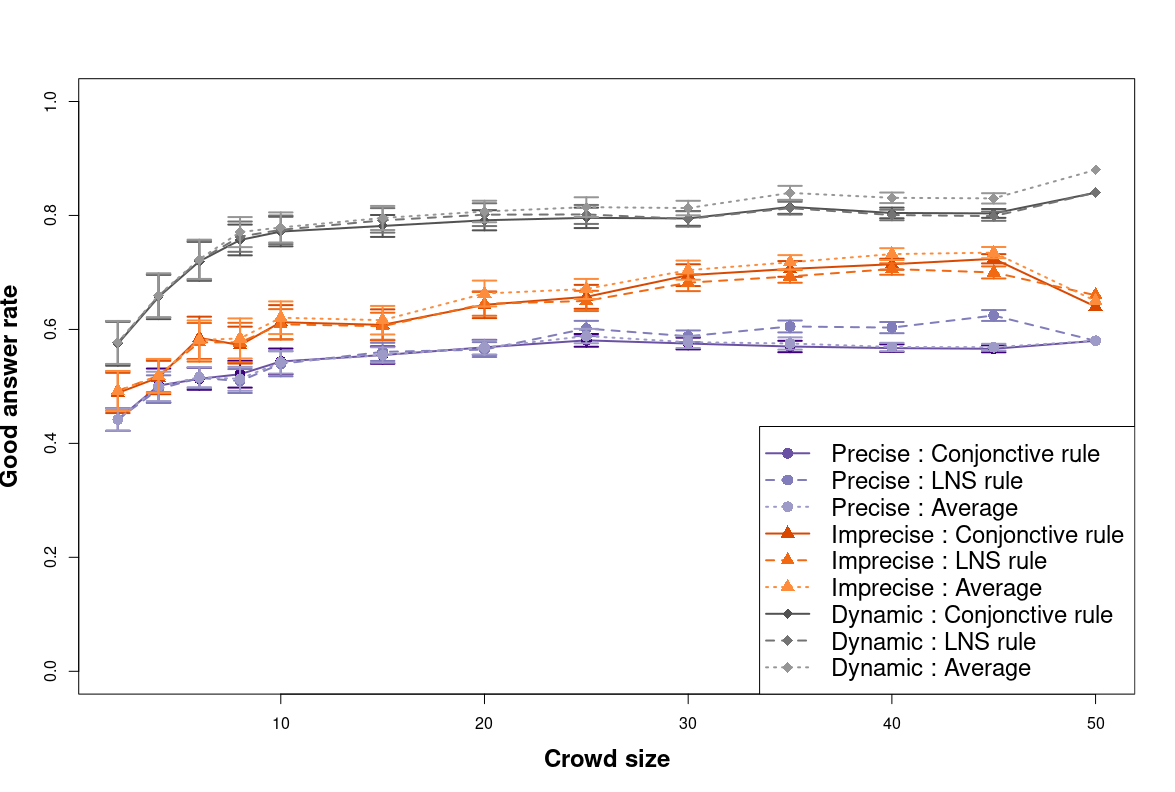}
    \caption{Comparison of operators: Conjunctive, LNS and mean for the campaigns {\bf 10\_birds\_precise, 10\_birds\_imprecise et 10\_birds\_dynamic}.}
    \label{fig:CRR_Foule_Op_10}
\end{figure}
Figure~\ref{fig:CRR_Foule_Op_10} is realized in a similar way to figure \ref{fig:CRR_Foule_Op_multi}, only $n \leq 50$ because this is the maximum crowd size for these campaigns.

Figure \ref{fig:CRR_Foule_Op_multi} shows that the gap between operators widens as the size of the crowd increases.
Figure \ref{fig:CRR_Foule_Op_10} shows close results between the different operators applied to the same dataset for a crowd size below 30 contributors.
Overall, according to these two figures, imprecise campaigns perform better than precise ones.
Moreover, the mean operator seems the most relevant of the three because for figure \ref{fig:CRR_Foule_Op_multi}, the mean offers the best results for the imprecise campaign.
For the precise campaign, it is the second best performing operator after the conjunctive rule.
In addition, on figure \ref{fig:CRR_Foule_Op_10} the CRRs of the mean are constantly the highest for campaigns with imprecise data.
For the campaign 10\_birds\_precise, the LNS rule shows the best results, but the confidence intervals of the LNS rule and the mean intersect up to a crowd size of $n=35$ contributors. 

After having established a comparison of combination operators we observe that the mean remains the method offering the highest CRR most commonly, so it is this operator that we use for the aggregation of the data by MONITOR in order to compare the model to the MV and EM.
\begin{table}[]
    \centering
    \begin{tabular}{|c|c|c|c|c|}
        \hline
        Campaign & $\alpha_I$ & $\alpha_R$ & $\alpha_C$ & $\alpha_A$ \\
        \hline
        multi\_birds\_precise & 0 & 2 & 6 & 2 \\
        \hline
        multi\_birds\_imprecise & 1 & 2 & 6 & 1 \\
        \hline
        10\_birds\_precise & 0 & 2 & 6 & 2 \\
        \hline
        10\_birds\_imprecise & 1 & 2 & 6 & 1 \\
        \hline
        10\_birds\_dynamics & 2 & 7 & 1 & 0 \\
        \hline
    \end{tabular}
    \caption{Coefficients used for the estimation of profiles by MONITOR for the different campaigns.}
    \label{tab:Coeff_profil_agg}
\end{table}
The coefficients used to estimate the profiles for the different campaigns are given in table \ref{tab:Coeff_profil_agg}.

In order to identify the best values of $\alpha_\mathcal{XP}$, the datasets are again divided into two to have a training dataset and a test dataset.
We consider that the Expert has better knowledge of the task domain and therefore more credit should be given to his answers. 
Thus for this category we test the values \linebreak $\alpha_{Expert} \in [0.5,1.0]$.
The Good contributor has less knowledge than the Expert that's why the $\alpha_{Good}$ values are tested on a more restricted interval $[0.5,0.85]$.
The average contributor, although willing to perform the task, lacks of capacity, so his answers can be further impaired and the interval used for the tests is $[0.2,0.7]$.
Finally the Bad contributor has random answers, so the employer cannot trust him and the values of $\alpha_{Bad}$ tested are included in the interval $[0.0,0.2]$.
The values retained for each campaign after learning are given in table \ref{tab:alpha_xp}.
\begin{table}[t]
    \centering
    \begin{tabular}{|c|c|c|c|c|}
        \hline
        Data & $\alpha_{Expert}$ & $\alpha_{Good}$ & $\alpha_{Average}$ &  $\alpha_{Bad}$\\
        \hline
        multi\_birds\_precise & 1 & 0.85 & 0.40 & 0.20 \\
        multi\_birds\_imprecise & 1 & 0.85 & 0.5 & 0.20 \\
        10\_birds\_precise & 1 & 0.5 & 0.2 & 0.05 \\
        10\_birds\_imprecise & 1 & 0.5 & 0.2 & 0.05 \\
        10\_birds\_dynamic & 1 & 0.60 & 0.3 & 0.15 \\
        \hline
    \end{tabular}
    \caption{$\alpha_\mathcal{XP}$ values used after learning}
    \label{tab:alpha_xp}
\end{table}
The values of table \ref{tab:alpha_xp} are such that:
$$\alpha_{Expert} > \alpha_{Good} > \alpha_{Average} \geq \alpha_{Bad}$$
In our tests, it happened that this ordering was not respected for equal correct response rates.
But there was always a set of values that respected this ordering for the best response rates.

\begin{figure}[t]
    \centering
    \includegraphics[height=8cm]{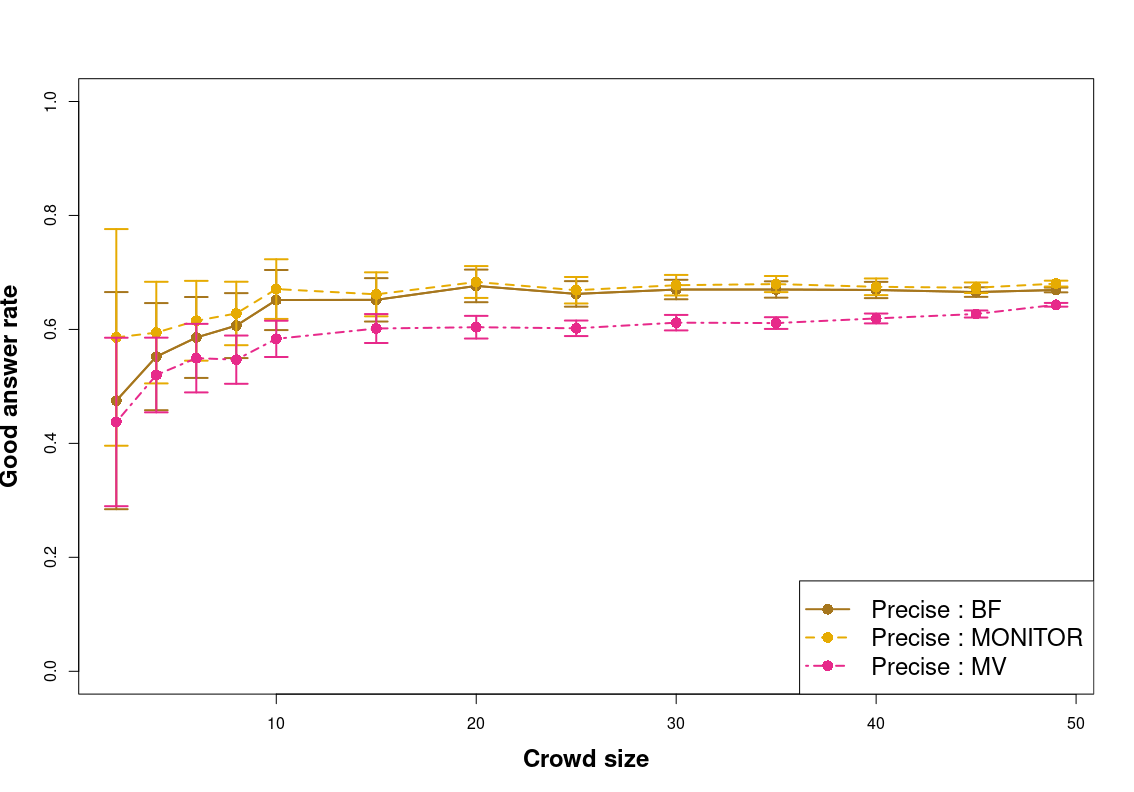}
    \caption{Comparison of the belief functions to the MV for the campaign {\bf multi\_birds\_precise}.}
    \label{fig:multi_precis_BF_MV}
\end{figure}

\begin{figure}[t]
    \centering
    \includegraphics[height=8cm]{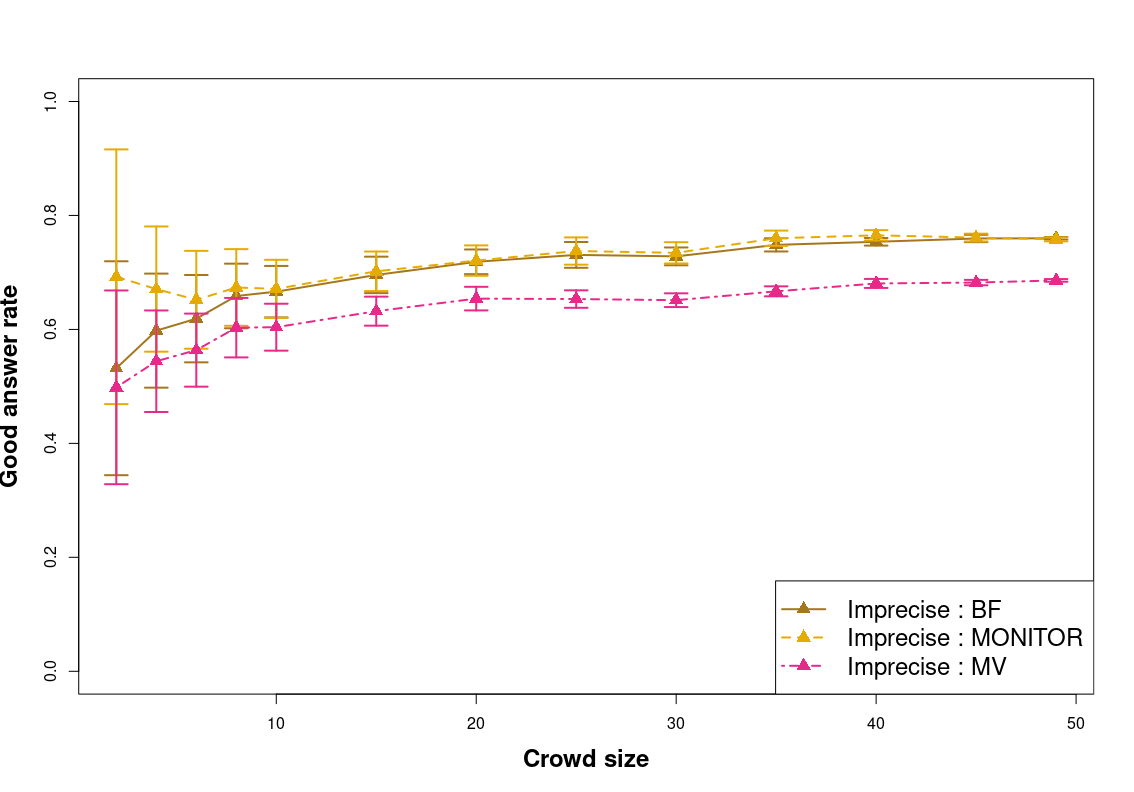}
    \caption{Comparison of the belief functions to the MV for the campaign {\bf multi\_birds\_imprecise}.}
    \label{fig:multi_imprecis_BF_MV}
\end{figure}

For the campaigns multi\_birds\_precise and multi\_birds\_imprecise, the answer set changes from question to question and each bird species pictured is unique in the database.
Since there is no repetition in the expected bird names, it is not possible to construct the confusion matrix required by EM. Thus for these two datasets the comparison is performed exclusively with the MV.
Figures \ref{fig:multi_precis_BF_MV} and \ref{fig:multi_imprecis_BF_MV} present the comparison between the MV, the mean of the mass functions made with a discounting coefficient $\alpha_\mathcal{XP}=0.9$ for all contributor's profiles and the aggregation performed by MONITOR.
The combination made by MONITOR uses the discounting coefficients given in table   \ref{tab:alpha_xp}.
For figures \ref{fig:multi_precis_BF_MV} and \ref{fig:multi_imprecis_BF_MV}, the CRRs are calculated for different crowd sizes $n$.
Thus, $n$ contributors are randomly selected from the set of contributors dedicated to the test data and their responses are aggregated.
This selection and aggregation method is performed 50 times for each $n$ crowd size and the CRRs are averaged to obtain the figure curves and their 95\% confidence intervals.
For both figures, the CRRs are increasing with the size $n$ of the crowd.
The CRR of MONITOR is slightly higher than that of the belief functions identically weakened by $\alpha=0.9$. Overall, the belief functions perform better than the MV.\\

\begin{figure}[t]
    \centering
    \includegraphics[height=8cm]{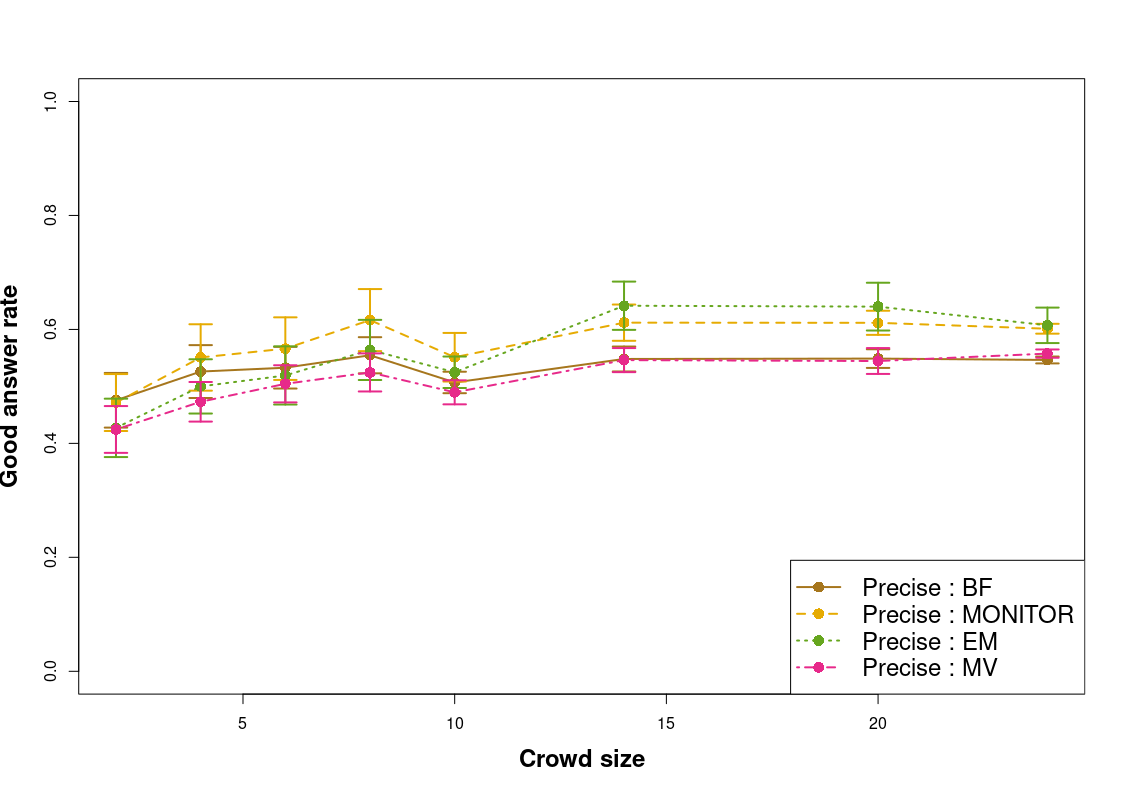}
    \caption{Comparison of aggregation methods for the campaign {\bf 10\_birds\_precise}.}
    \label{fig:10_precis_BF_EM_MV}
\end{figure}

\begin{figure}[t]
    \centering
    \includegraphics[height=8cm]{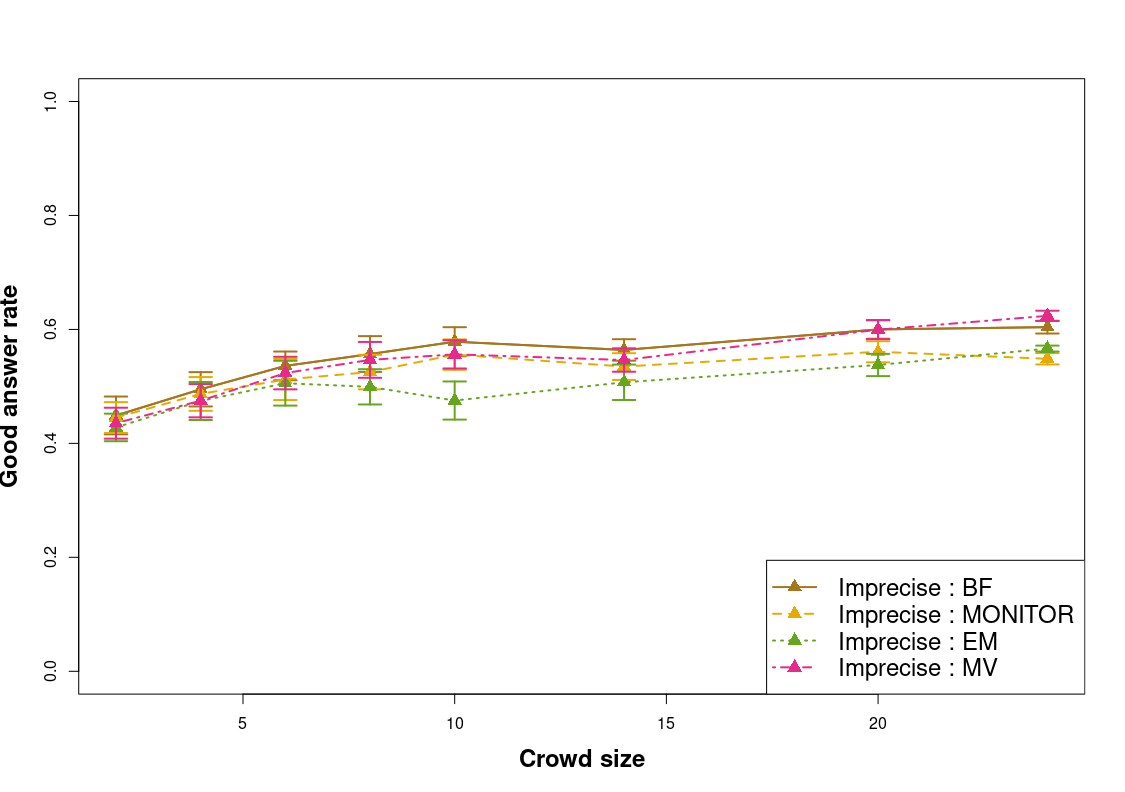}
    \caption{Comparison of aggregation methods for the campaign {\bf 10\_birds\_imprecise}.}
    \label{fig:10_imprecis_BF_EM_MV}
\end{figure}

\begin{figure}[t]
    \centering
    \includegraphics[height=8cm]{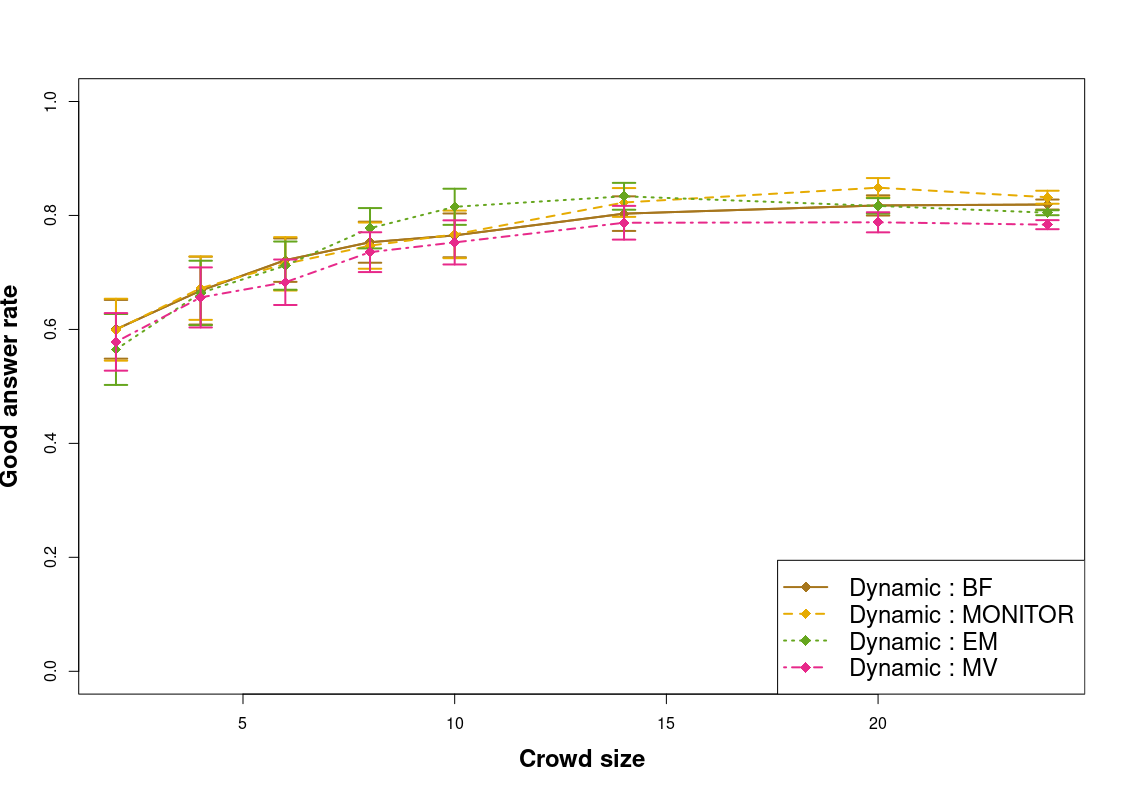}
    \caption{Comparison of aggregation methods for the campaign {\bf 10\_birds\_dynamic}.}
    \label{fig:10_dynamique_BF_EM_MV}
\end{figure}

The use of the same 10 bird species throughout the campaigns 10\_birds\_precise, 10\_birds\_imprecise and 10\_birds\_dynamic allows this time a comparison of the MV and the belief functions with EM.
For the data 10\_birds\_dynamic, the possible iteration on the answer is not used, only the first contribution filled in is exploited.
To obtain figures \ref{fig:10_precis_BF_EM_MV}, \ref{fig:10_imprecis_BF_EM_MV} and \ref{fig:10_dynamique_BF_EM_MV} the contributions of the test sets are aggregated 25 times for an $n$ crowd size. 

Figure \ref{fig:10_precis_BF_EM_MV} shows that MONITOR and EM perform better than MV and the average of the mass functions.
For 10 contributors or less MONITOR offers the best results and for a larger crowd EM performs better, but whatever the size of the crowd the confidence intervals of MONITOR and EM overlap showing a closeness of results. 
For figure \ref{fig:10_imprecis_BF_EM_MV}, belief functions aggregated by the mean perform better overall than other aggregation methods, EM and MONITOR have lower CRRs than MV.
According to figure \ref{fig:10_dynamique_BF_EM_MV} EM and MONITOR are again the two approaches giving the best rates of correct answers with overlapping confidence intervals.
The MV gives the worst results. 


\section{Conclusion}
\label{sec:ccl}

Crowdsourcing is characterized by the outsourcing of tasks to a crowd of contributors on dedicated platforms.
The crowd includes contributors with various profiles whose contributions do not all have the same relevance, which poses the problem of estimating the contributor's profile and also of aggregating the responses.
Indeed, the collected data present imperfections related to human contributions.
The methods currently used (MV and EM), whether for estimating the profile or aggregating the responses, have their limitations.
For MV there is no profile estimation and all answers have the same weight in the aggregation.
For EM, the estimation of the contributor's profile only takes into account his qualification for the task and not his seriousness in performing it.
In order to overcome these problems, we defined MONITOR for the estimation of the contributor's profile based on his qualification and his behavior.
The proposed model also allows the aggregation of uncertain and imprecise answers while taking into account the profile of the contributor.

In order to carry out our tests on real data we have conducted five crowdsourcing campaigns which consist in annotating pictures of birds.
For all the campaigns, the contributor is asked to indicate his certainty in his answer.
For the first two campaigns, the proposed answers changed for each question, for one it is required from the contributor a precise answer, 
for the second, the contributor can be imprecise and select several answers. 
For the other three campaigns, only ten bird species are to be identified, and the ten names are proposed for each question.
For the third campaign 
the contributor must select a single name, while for the fourth and fifth campaigns 
he can be imprecise and choose up to 5 bird names. 
These data allowed us to conduct experiments on the elements that make up the profile, on the profile itself and on the aggregation of responses.

We compared MONITOR's estimate of precision to the degree of \cite{rjab16} from which it was derived.
Our modeling of contributor's precision is more in line with the actual average contributor's precision than the degree proposed by the authors.
For contributor's reflection verification, we compare our computational method using belief functions to a statistical contributor's selection approach used by \cite{komarov13}.
The authors' approach is limited by the statistical distribution of contributor's response times, which can be a source of error.
MONITOR provides a relevant estimation of the reflection as a function of the contributor's response time.
However, reflection alone cannot be considered to exclude a contributor because a non-reflective contributor can be a spammer as well as an expert.
It is thus essential to consider reflection in conjunction with attention in order to determine the behavior of the contributor.

Experiments performed for the profile estimation by MONITOR compared to the estimation of \cite{rjab16} and EM show that MONITOR provides a better rate of good profile classification than \cite{rjab16} while being less expensive in computation time.
However, an estimate of the contributor's profile using EM is still more efficient than the estimates obtained by MONITOR.
Nevertheless, it is not always possible to apply the EM algorithm to the collected data, especially in the case where the set of answers proposed to the contributor changes from one question to another.
MONITOR on the other hand is not impacted by this issue and is applicable to any type of imprecise and uncertain data.

For the data of the first and second campaign, it is note possible to use EM.
The results obtained on these data show that discounting the contributions according to the profiles estimated by MONITOR offer better good response rates than a common discounting for the whole data.
For both campaigns, the belief functions offer better results than the MV traditionally used in crowdsourcing platforms.
For campaigns with 10 recurrent bird species to be identified, the method giving the best correct response rates changes from campaign to campaign between MV, EM and MONITOR.
This shows that the definition of the campaign has a strong impact on the contributions collected and the aggregation method to be used afterwards.
For campaigns similar to those with a multitude of birds, where a learning by the contributor is more complex, the most appropriate method to use is MONITOR.
In the case of redundant questions, it is possible to use EM as MONITOR.
The model can be further optimized for dynamic use during the crowdsourcing campaign, this is the subject of our research perspectives.

Currently MONITOR is applied on the data collected at the end of the crowdsourcing campaign. In our future work we want to estimate the profile of the contributor during the campaign thanks to MONITOR. This research perspective raises the issue of using belief functions in a dynamic context.


\section{Acknowledgements}

We are grateful to the departmental council of Côtes-d'Armor and the ANR Headwork project for funding this work.

\bibliography{biblio.bib}

\end{document}